\documentclass{article}

\usepackage{microtype}
\usepackage{graphicx}
\usepackage{subfigure}
\usepackage{booktabs} %

\usepackage{hyperref}
\usepackage[dvipsnames]{xcolor}%

\usepackage[arxiv]{icml2025}
\usepackage{amsmath}
\usepackage{amssymb}
\usepackage{mathtools}
\usepackage{amsthm}

\usepackage[capitalize]{cleveref}

\icmltitlerunning{Robust Autonomy Emerges from Self-Play}

\usepackage{graphicx}%
\usepackage{multirow}%
\usepackage{amsmath,amssymb,amsfonts}%
\usepackage{amsthm}%
\usepackage{mathrsfs}%
\usepackage[title]{appendix}%
\usepackage{textcomp}%
\usepackage{manyfoot}%
\usepackage{booktabs}%
\usepackage{algorithm}%
\usepackage{algorithmicx}%
\usepackage{algpseudocode}%
\usepackage{listings}%
\usepackage{verbatim}%
\usepackage{setspace}%
\usepackage{placeins}%

\usepackage{hyperref}
\usepackage{siunitx} %

\usepackage{colortbl}
\usepackage{xspace}
\usepackage{xparse}
\usepackage{bbm}
\usepackage{booktabs}

\usepackage{tabularx}

\newcommand\clearrow{\global\let\rowmac\relax}
\clearrow

\usepackage{subcaption}

\usepackage{bold-extra}  %

\NewDocumentCommand{\todo}{om}{\textcolor{red}{Todo\IfValueT{#1}{ (#1)}: #2}}
\newcommand\ours{\textsc{Gigaflow}\xspace}

\newcommand\mypara[1]{\noindent\textbf{#1}}

\newcommand{\mynum}[1]{%
    \num[round-precision=2,round-mode=figures,scientific-notation=true]{#1}%
}
\newcommand{\eg}{\text{e.g.,}\xspace}

\begin{document}

\twocolumn[
\icmltitle{Robust Autonomy Emerges from Self-Play}

\icmlsetsymbol{equal}{*}

\begin{icmlauthorlist}

\icmlauthor{Marco Cusumano-Towner}{equal,apple}
\icmlauthor{David Hafner}{equal,apple}
\icmlauthor{Alex Hertzberg}{equal,apple}
\icmlauthor{Brody Huval}{equal,apple}
\icmlauthor{Aleksei Petrenko}{equal,apple}
\icmlauthor{Eugene Vinitsky}{equal,apple}
\icmlauthor{Erik Wijmans}{equal,apple}
\icmlauthor{Taylor Killian}{apple}
\icmlauthor{Stuart Bowers}{apple}
\icmlauthor{Ozan Sener}{apple}

\icmlauthor{Philipp Kr\"ahenb\"uhl}{apple}
\icmlauthor{Vladlen Koltun}{apple}

\end{icmlauthorlist}

\icmlaffiliation{apple}{Apple}

\icmlcorrespondingauthor{Philipp Kr\"ahenb\"uhl}{philkr@apple.com}

\icmlkeywords{Machine Learning, ICML}

\vskip 0.3in
]

\printAffiliationsAndNotice{\textsuperscript{*}Equal contribution, alphabetical order.} %

\def\acite#1{}
\begin{abstract}

    Self-play has powered breakthroughs in two-player and multi-player games.
    Here we show that self-play is a surprisingly effective strategy in another domain. We show that robust and naturalistic driving emerges entirely from self-play in simulation at unprecedented scale -- 1.6~billion~km of driving.
    This is enabled by \ours, a batched simulator that can synthesize and train on 42 years of subjective driving experience per hour on a single 8-GPU node.
    The resulting policy achieves state-of-the-art performance on three independent autonomous driving benchmarks.
    The policy outperforms the prior state of the art when tested on recorded real-world scenarios, amidst human drivers, without ever seeing human data during training.
    The policy is realistic when assessed against human references and achieves unprecedented robustness, averaging 17.5 years of continuous driving between incidents in simulation.
\end{abstract}

\section{Introduction}

\begin{figure*}[ht!]
  \centering
    \includegraphics[width=0.9\linewidth,page=1]{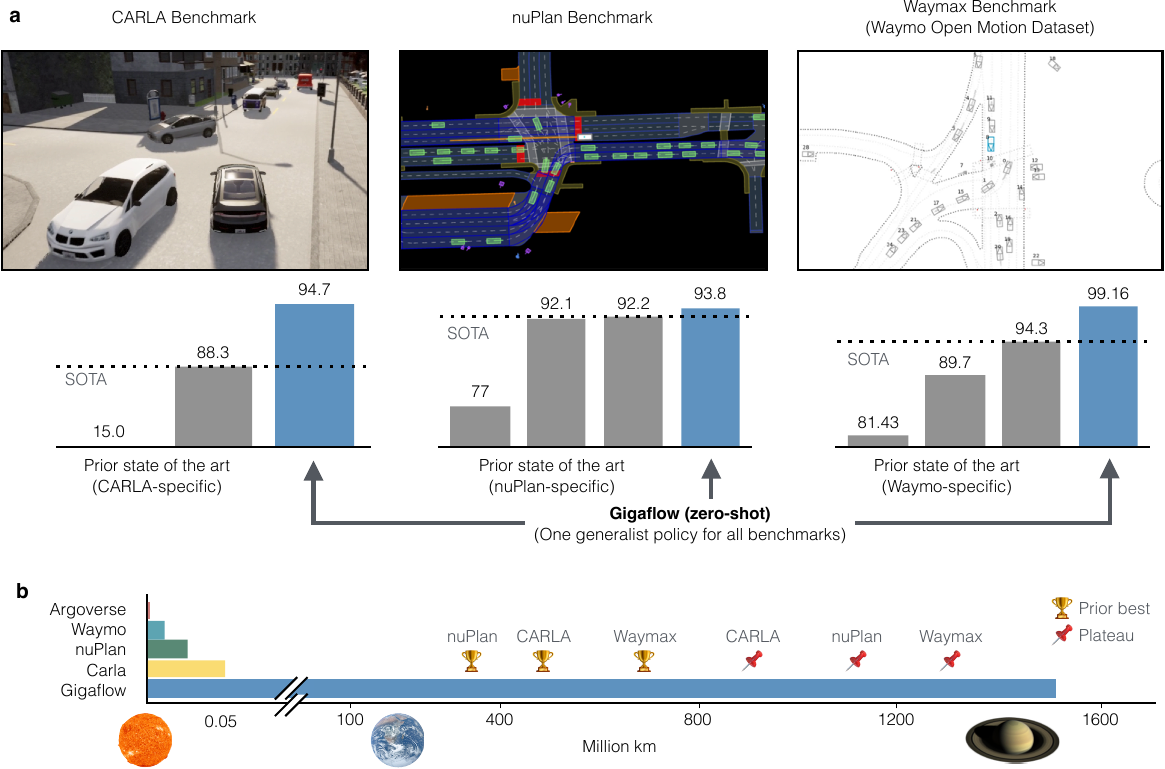}
    \caption{
    \textbf{Self-play reinforcement learning yields a generalist policy}.
    \textbf{a}, A single \ours agent outperforms the best dataset-specific specialists across leading benchmarks.
    \ours is evaluated zero-shot without training on a single benchmark, while dataset-specific specialists train on benchmark-specific datasets.
    Each benchmark includes different maps, scenarios, and evaluation metrics.
    \textbf{b}, \ours enables cost-effective training of policies via self-play on a massive scale.
    Our largest policies drive over 1.6 billion km during training, more than the distance from the Sun to Saturn and orders of magnitude farther than prior datasets or simulations.
    At this scale, self-play yields a generalist policy.
    Dashed lines indicate points at which the performance of our single generalist policy passes the prior state of the art on each benchmark (`prior best') and points at which the performance on each benchmark plateaus (`plateau').
    }
    \label{fig:teaser}
\end{figure*}

Self-play has been an effective strategy for training policies for board games, card games, 3D multiplayer games, real-time strategy games, robotic manipulation, and even bioengineering~\cite{Silver2017,Silver2018,Jaderberg2019,Berner2019,BrownSandholm2019,Vinyals2019,Plappert2021,Perolat2022,Wang2024:NMI}. In this work, we demonstrate the effectiveness of self-play in another domain. We show that simulated self-play yields naturalistic and robust driving policies, while using only a minimalistic reward function and never seeing human data during training.

We demonstrate that qualitatively new levels of realism and robustness emerge when self-play training is taken to unprecedented scale~-- orders of magnitude beyond prior experiments~\cite{Feng2023,Zhang2023:CoRL}.
This discovery is enabled by \ours, a batched simulator architected from the ground up for self-play reinforcement learning on a massive scale.
\ours is capable of simulating and learning from $4.4$ billion state transitions (7.2 million km of driving, or 42 years of continuous driving experience) per hour on a single 8-GPU node.
It simulates urban environments with up to 150 densely interacting traffic participants \num{360000} times faster than real time at a cost of under \$5 per million km driven (based on public cloud rates).
A full training run simulates over one trillion state transitions, 1.6 billion km driven, or \num{9500} years of subjective driving experience, and completes in under 10 days one 8-GPU node.

We use \ours to train a parameterized family of driving policies.
The parameters specify the type of traffic participant controlled by the policy (passenger vehicle, large truck, bicyclist, or even a pedestrian) and the driving style (\eg, aggressive vs.\ cautious).
These parameters can be modified at test time with no additional training~\cite{DosovitskiyKoltun2017}, such that a single trained policy can be used to control a variety of traffic participants, with a variety of behavioral styles.
During training, this parameterized policy architecture enables all simulated traffic participants to be collecting experience in parallel, all flowing through a single neural network.
This supports self-play simulations where more than a hundred agents are all controlled by a single neural network, which is learning from all of their experiences, yet the agents exhibit diverse outward manifestations (truck vs.\ bicycle), functional characteristics (turning radius), and behavioral styles (adherence to traffic laws).

The result is a robust and naturalistic driving policy that achieves state-of-the-art performance when tested in recorded real-world scenarios, amidst recorded human drivers, without ever seeing human data during training.
We test the \ours policy in three leading independent third-party benchmarks: CARLA~\cite{dosovitskiy2017}, nuPlan~\cite{caesar2022nuplan}, and the Waymo Open Motion Dataset~\cite{ettinger2021large} (through the Waymax simulator~\cite{gulino2023waymax}).
State-of-the-art performance on each benchmark was previously achieved by specialist agents that were trained specifically for that benchmark, commonly using benchmark-specific datasets.
In contrast, we outperform the prior state of the art on all benchmarks with a single policy (\cref{fig:teaser}) that was trained entirely via self-play, using none of the provided datasets for training.

The behaviors exhibited by the \ours policy are naturalistic
despite never seeing human data during training.
The trained policy exhibits long-horizon planning without any dedicated planning or search modules, can deal with heavily contentious traffic scenarios,
is quantitatively realistic when assessed against human references~\cite{montali2023waymo}, and exhibits unprecedented robustness, averaging over 3 million km (or 17.5 years of continuous driving) between incidents in simulation.

\begin{figure*}[ht!]
    \centering
    \includegraphics[width=0.9\linewidth,page=1]{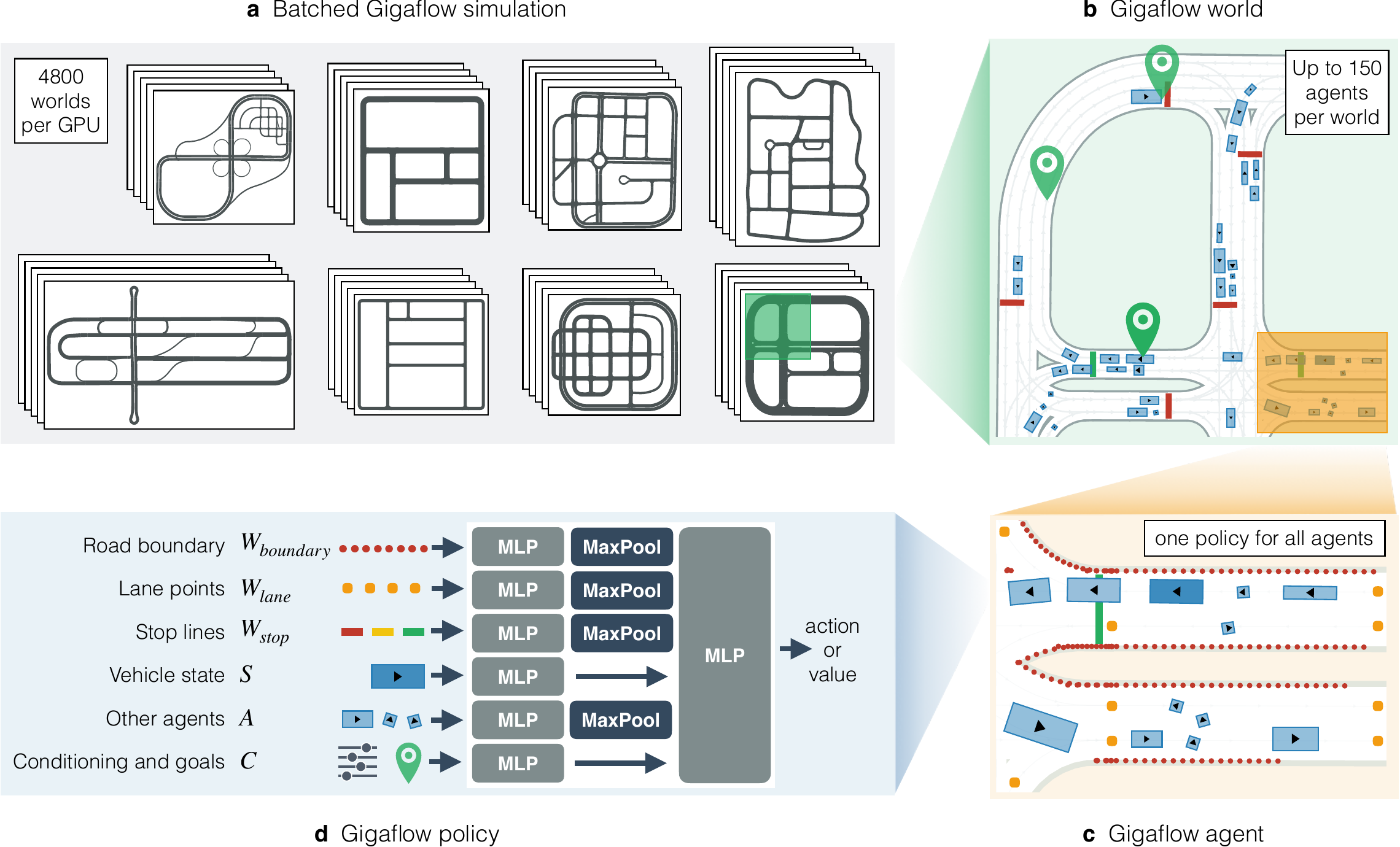}
    \vspace{-2mm}
    \caption{
    \textbf{Overview of \ours.}
    \textbf{a}, \ours simulates tens of thousands of worlds with millions of agents in a massively parallel self-play reinforcement learning setup.
    \textbf{b}, Each world tasks agents to navigate to goals on a map without collisions.
    \textbf{c}, Each agent optimizes its own performance, given a set of local observations.
    \textbf{d}, All agents use a compact shared policy network.
    }
    \label{fig:technical}
    \vspace{-3mm}
\end{figure*}

\section{\ours}
\label{sec:technical}

The goal of \ours is to train a \emph{generalist policy} $\pi(a | W, S, A, C)$ in simulation (\cref{fig:technical}d).
The policy observes the static world $W$, its own state $S$, and other dynamic agents $A$ to produce an action $a$ (\cref{fig:technical}c).
A conditioning parameter $C$ modulates the policy's behavior.
Learning this generalist policy requires careful modeling of two core concepts: uncertainty and other-agent behaviors.

\mypara{Uncertainty} in driving stems from partial or incomplete observations.
The driver is generally unaware of the goals and intentions of other agents, or even their exact location, speed, or acceleration. Objects or parts of the static world may be hidden or occluded.
Real-world sensors often introduce noise.
\ours models uncertainty directly through noise on the state $S$, noise in the state transitions, stochasticity in the dynamic agents, and partial observability on dynamic agents $A$ and the static world $W$. \ours agents observe the positions and speeds of nearby agents but not their acceleration, and crucially, neither their goals nor conditioning.

\mypara {Modeling agent behaviors} is a particularly impactful and complex aspect of driving.
Prior work approached behavior models through hand-designed agents~\cite{kesting2010enhanced, kesting2007general, gulino2023waymax}, recorded and replayed data~\cite{gulino2023waymax, li2022metadrive, vinitsky2022nocturne}, or models of driving learned from data~\cite{ivanovic2019trajectron, nayakanti2023wayformer, wang2023multiverse, suo2021trafficsim, shi2022motion, xu2023bits, zhong2023guided}.
In contrast, in \ours, realistic and general driving behaviors for all traffic participants emerge via self-play reinforcement learning on a massive scale.

\subsection{\ours world}
The \ours world is simple:
we do not script scenarios, use human driving traces, or design delicate reward terms.
We show that simulation at massive scale makes up for much of this simplicity (\cref{fig:technical}).

Agents train on one of eight maps, randomly perturbed with rescaling, shears, flips and reflections. Total drivable lanes per map range from four to 40 km for a total of 136 km of road across the eight maps (\cref{fig:technical}a).
In each map, we spawn one to $N_a$ agents at random locations and orientations on the road and ask them to reach goal points sampled uniformly over the map.
This creates a world in which agents drive for long distances before reaching their destinations (\cref{fig:technical}b).
Agents are tasked with visiting a variable number of intermediate waypoints, requiring the ability to follow complex routes (see \cref{sec:pos_game} for details).

Dense traffic flows with diverse interactions emerge as agents navigate to their destinations.
As training progresses, we can observe agents executing zipper merges and tight maneuvers in traffic jams, managing congested roundabouts and uncontrolled intersections, resolving occasional gridlocks, and performing multi-point turns to reroute around accidents or obstructions.

\ours agents train fully in self-play.
All dynamic agents~-- vehicles, pedestrians, and cyclists~-- use the same single reactive parametric policy $\pi$; their behaviors are varied through conditioning $C$ (\cref{fig:technical}d).
The policy is aware of the dynamics of the agent it controls as part of the conditioning $C_\text{dynamics}$. The agent reward is a mixture of incentives to reach its goal, avoid collisions, drive centered and lane aligned, as well as penalties for running red lights or stop signs, and exceeding acceleration and jerk limits. The weights on each of these reward components are randomized per agent and provided as conditioning $C_\text{reward}$ to the agent (see \cref{sec:pos_game} for details).
This allows a single reactive policy $\pi$ to exhibit a wide range of behaviors. The result is a diverse training world where agents learn a continuum of driving styles: some drive cautiously, others are likely to run traffic lights, while a rare few are willing to drive against the flow of traffic.
Because the policy only observes the conditioning $C$ of the agent it controls, it must learn to be robust to the unpredictable behaviors of other drivers.

\label{sec:problem_formulation}

\subsection{\ours simulation and training}
\label{sec:scaling}

The \ours simulation and training framework is designed to optimize driving data collection and training throughput per unit of computation.
We simulate and learn from $4.4$ billion
state transitions per hour on a single 8-GPU node, rolling out urban commute simulations \num{360000} times faster than real time at the cost of under \$5 per million kilometers driven (based on public cloud rates).
These training rates require three core ingredients: a fast batched simulator~\cite{Shacklett2021,petrenko2021megaverse,shacklett23madrona}, a compact and expressive policy for fast inference and backpropagation, and a high-throughput training algorithm.

\mypara{\ours simulation.}
\ours simulates \num{38400}
environments in parallel across 8 GPUs with up to $N_a=150$ vehicles each (\cref{fig:technical}a).
Basic operations, such as policy inference and dynamics updates are batched across all agents.
Agent localization, collision checking, and observation construction rely on dedicated optimized data structures. Due to the large map sizes, we precompute and cache all map observations in a spatial hash and perform fast, GPU-based runtime lookup and retrieval. Agents perceive the map by observing sets of points sampled sparsely along drivable lanes $W_\text{lane}$, and densely along the nearby road edges for precise maneuvering $W_\text{boundary}$ (Fig. \ref{fig:technical}c).

Beyond map features, agents get observations of nearby traffic participants $A$ -- containing %
nearby vehicles' sizes, locations, orientations, and velocities -- and nearby stop lines and traffic lights $W_\text{stop}$.
\ours models static obstacles as immobile vehicles $A_\text{static}$.
To reduce memory, we do not store these observations in the rollout buffer, but calculate them on demand from stored world states.
See \cref{sec:sim_design} for a detailed description of the simulator.

\mypara{\ours policy.}
\ours can simulate diverse actor types, from pedestrians to heavy trucks, by parameterizing a single unified feed-forward policy (\cref{fig:technical}d).
The decision to use the same underlying neural network policy for all traffic participants significantly impacts the overall throughput: we need only a single (batched) forward pass per simulation step to calculate actions for all agents.
The policy resembles a Deep Sets architecture~\cite{Zaheer2017} and is invariant to permutation \text{w.r.t.} each observation type.
Critically, the entire trainable artifact is relatively compact at six million parameters.
 On an 8-GPU A100 node, the policy allows inference throughput of $7.4$ million
 decisions per second during experience collection at a batch size of $2.6$ million,
 and eight gradient updates per second in the training phase with a batch size of \num{256000}.
 See \cref{sec:policy_arch} for more details.

\mypara{\ours training.}
We train the \ours policy using Proximal Policy Optimization (PPO)~\cite{schulman2017ppo}. One of the main challenges associated with autonomous driving is the inherent imbalance in the data distribution. As training progresses, the on-policy data is dominated by ordinary traffic configurations, such as orderly driving in a straight line between intersections.
The critic is often able to accurately predict the returns for such trajectories, resulting in a large portion of samples with near-zero advantage~\cite{greensmith2004variance} that consequently yield vanishingly small gradients.

We use a variant of Prioritized Experience Replay~\cite{schaul2016prioritized} that filters samples that have minimal impact on learning. The filtering is based on the absolute value of the estimated advantage. We filter up to 80\% of samples with low absolute advantage, which significantly increases learning throughput without sacrificing sample efficiency. Our approach, which we refer to as \emph{advantage filtering}, focuses training on the most informative state transitions, prioritizing learning from the underexplored tails of the data distribution where selected actions are measurably better or worse, and makes more efficient use of the data we generate.
See \cref{sec:training_algo} for more details.
\section{Zero-shot evaluation on driving benchmarks}
\label{sec:zeroshot}

We evaluate a trained \ours policy on the leading closed-loop driving benchmarks: CARLA~\cite{dosovitskiy2017}, nuPlan~\cite{caesar2022nuplan}, and the Waymo Open Motion Dataset~\cite{ettinger2021large} through the Waymax simulator~\cite{gulino2023waymax}.
These benchmarks encompass a wide range of actor behaviors, driving scenarios, maps, traffic densities, durations, and scoring methodologies.
The CARLA benchmark consists of routes with hand-designed scenarios based on the NHTSA pre-crash topology~\cite{wassim2007nhtsa}. It evaluates long distance driving (several minutes per 1--3\,km route). nuPlan and Waymax evaluate short distance driving (8--14 seconds per scenario, $<\!100$\,m) in scenarios derived from recorded real-world driving with the associated sensor data.

\mypara{A generalist \ours policy outperforms state-of-the-art specialists.}
For each benchmark, we compare to specialist state-of-the-art policies that are either trained~\cite{gulino2023waymax} or carefully hand-designed~\cite{chitta2023transfuser,jaeger2023hidden,dauner2023parting} to perform well on that specific benchmark.
In contrast, we use a single policy across all benchmarks. Our policy is trained purely in self-play and is evaluated zero-shot in each benchmark environment.
Without any fine-tuning, our policy surpasses the state of the art in CARLA, nuPlan, and Waymax (\cref{fig:teaser} with details in \cref{tab:carla_results,tab:nuplan_results,tab:waymax_results} and \cref{sec:supplement_eval}).
This demonstrates robust driving with strong generalization.
Our self-play policy outperforms the state of the art on real driving traces with human traffic participants, without ever seeing human data during training.

\mypara{\ours policy generalizes to diverse actor behaviors.}
The benchmarks implement a diverse set of environment actors.
CARLA uses reactive rules-based vehicles with lane-changing capabilities, combined with events triggered by the driver's behavior (\eg a pedestrian
that darts suddenly in front of the driver).
The actors in nuPlan and Waymax are controlled by different variants of the Intelligent Driver Model~\cite{Treiber_2000}.
Vehicles in nuPlan follow the lane center line, whereas vehicles in Waymax follow the paths of logged human drivers.
The \ours policy exhibits robust driving amongst all of these actor types.

\mypara{\ours policy generalizes to diverse maps and driving situations.}
\ours trains on variants of synthetic maps with closed road networks~\cite{dosovitskiy2017}, but generalizes to the real-world maps in nuPlan~\cite{caesar2022nuplan} and in the Waymo Open Motion Dataset (WOMD)~\cite{ettinger2021large}.
The WOMD maps are small, with incomplete road networks constructed from logs of instrumented vehicles in several US cities.
The nuPlan benchmark is based on driving logs of human drivers in locales with both right-handed and left-handed driving~\cite{Caesar_2020_CVPR}; it contains larger maps that encompass the entire testing area of the vehicle.
Both nuPlan and WOMD scenarios include merges, unprotected turns, and interactions with pedestrians and cyclists~\cite{ettinger2021large}.
The \ours policy achieves state-of-the-art results in these benchmarks without any training on recorded driving logs or any human-designed scenarios.

\mypara{\ours policy generalizes to real-world observation noise.}
Both Waymax and nuPlan construct observations, maps, and other actors with auto-labeling tools from real-world perception data.
This brings occlusion, incorrect or missing traffic-light states, and obstacles revealed at the last moment.
Despite the minimalistic noise modeling in \ours, the \ours policy generalizes zero-shot to these conditions.

\mypara{\ours policy is state-of-the-art according to multiple scoring methodologies.}
Each benchmark brings its own definition of `good driving'. Those definitions are distinct and sometimes contradictory. For example, running a red light in CARLA incurs nearly the same penalty as colliding with another vehicle. Yet the same action can be advantageous in nuPlan, where red light violations are ignored by the scoring criteria, hard braking causes comfort penalties, and forward progress is strongly rewarded.
Despite such variations, the single generalist \ours driver outperforms specialist policies optimized for individual benchmark scores.

\mypara{The \ours policy approaches the ceiling of benchmark performance.}
The vast majority of the infractions sustained by the \ours policy during testing on the benchmarks can be attributed to limitations of the benchmarks. For instance, $20\%$ of the reported infractions in CARLA are caused by pedestrians or cyclists darting from the sidewalk into the roadway without reacting to the evasive maneuver of the driver or other traffic participants. Preventing such collisions would require drastic overfitting to this type of scenario~\cite{jaeger2023hidden}. Other exemplary limitations are gridlocks caused by CARLA-controlled traffic ($33\%$ of all infractions) or fuzzy stop sign and red light checks ($16\%$ of all infractions).

In nuPlan our policy sustains $15$ collisions in $\num{1118}$ scenarios. We analyzed each of them. Nine are unavoidable due to invalid initialization or sensor noise (agents appearing inside the vehicle's bounding box).
Four are caused by non-reactive pedestrian agents walking into the vehicle while the vehicle was stopped or in an evasive maneuver.
Two collisions are due to traffic light violations of other agents.

In Waymax our policy sustains $187$ collisions in $\num{44097}$ scenarios. We again analyzed each of them. $55.6\%$ were caused by unavoidable IDM agent behavior~\cite{Treiber_2000} of the traffic participants controlled by the benchmark, such as swerving directly into the ego vehicle. $41.7\%$ were caused by initialization in a state of collision, typically with a pedestrian. $2.7\%$ (i.e. five scenarios) were considered at-fault and avoidable by the \ours policy. Of the at-fault collisions, there were additional contributing factors such as perception issues or aggressive and spurious IDM behaviors. One example is when the \ours policy seeks to avoid a rear-end collision with an IDM agent approaching from behind at high speed.

We include videos of all reported infractions in the supplementary material.

\begin{figure*}[ht!]
    \centering
        \includegraphics[width=1\linewidth]{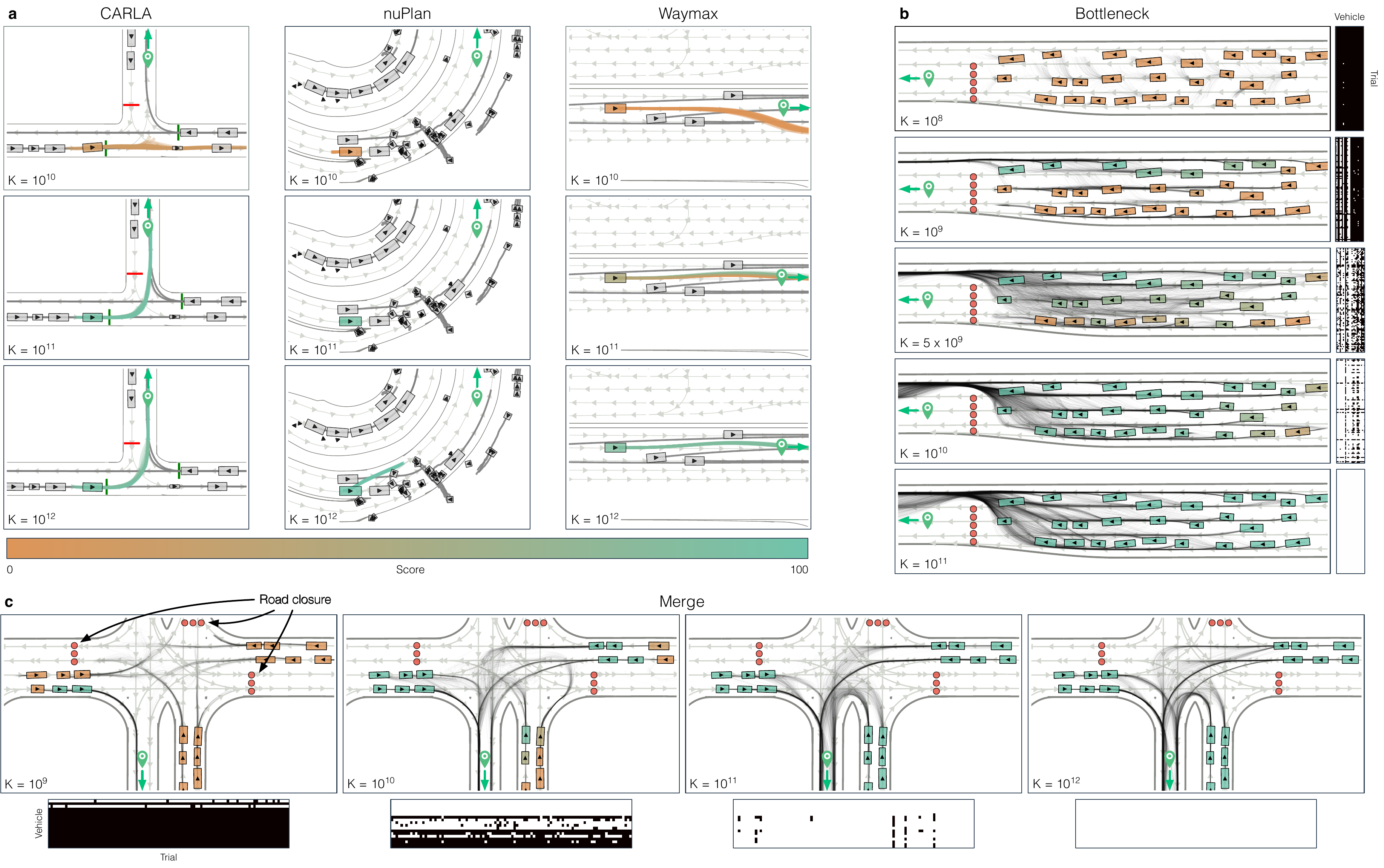}
    \vspace{-2mm}
    \caption{
\textbf{Evolution of the policy during training.}
\textbf{a},
Zero-shot performance on example benchmark scenarios with increasing number of state transitions ($K$) seen during training. From top to bottom: $K = 10^{10}$ (3 million km), $K = 10^{11}$ (90 million km), and $K = 10^{12}$ (1600 million km). The color of the controlled vehicle and its trajectories indicate the benchmark score.
\textbf{b},
Performance of the \ours policy on a diagnostic scenario where a static obstruction (red bounding boxes) forces a merge of several highway lanes into one.
All agents are controlled by the same policy.
Trajectories from 100 independent rollouts are shown.
Agent colors indicate probability of success (moving past the bottleneck).
Binary matrices on the right indicate the success of each actor in each rollout (white indicates success).
After $K = 10^8$ state transitions, the agents drive out of their lanes and crash.
After $K = 10^9$, the agents drive forward, avoid collision, and sometimes successfully change lanes, but do not successfully merge across multiple lanes (only agents on the right succeed).
After $K = 5 \times 10^9$, merging ability emerges but agents in the leftmost lane usually fail.
After $K = 10^{10}$, agents sometimes merge from the left lane, but not reliably.
After $K = 10^{11}$, all agents reliably succeed with no incidents.
\textbf{c}, Performance of the \ours policy in a diagnostic scenario where road closures require three traffic flows to merge into one without the aid of traffic lights. As training progresses, the policy goes from only succeeding at the right turn to successfully completing unprotected left turns and u-turns.
After $K = 10^{12}$ steps of training, all agents reliably succeed. Binary matrices show the worst 100 samples from 1000.
}
    \label{fig:training-analysis}
    \vspace{-5mm}
\end{figure*}

\section{Analysis}

\begin{figure*}[ht!]
    \centering
    \includegraphics[width=\linewidth]{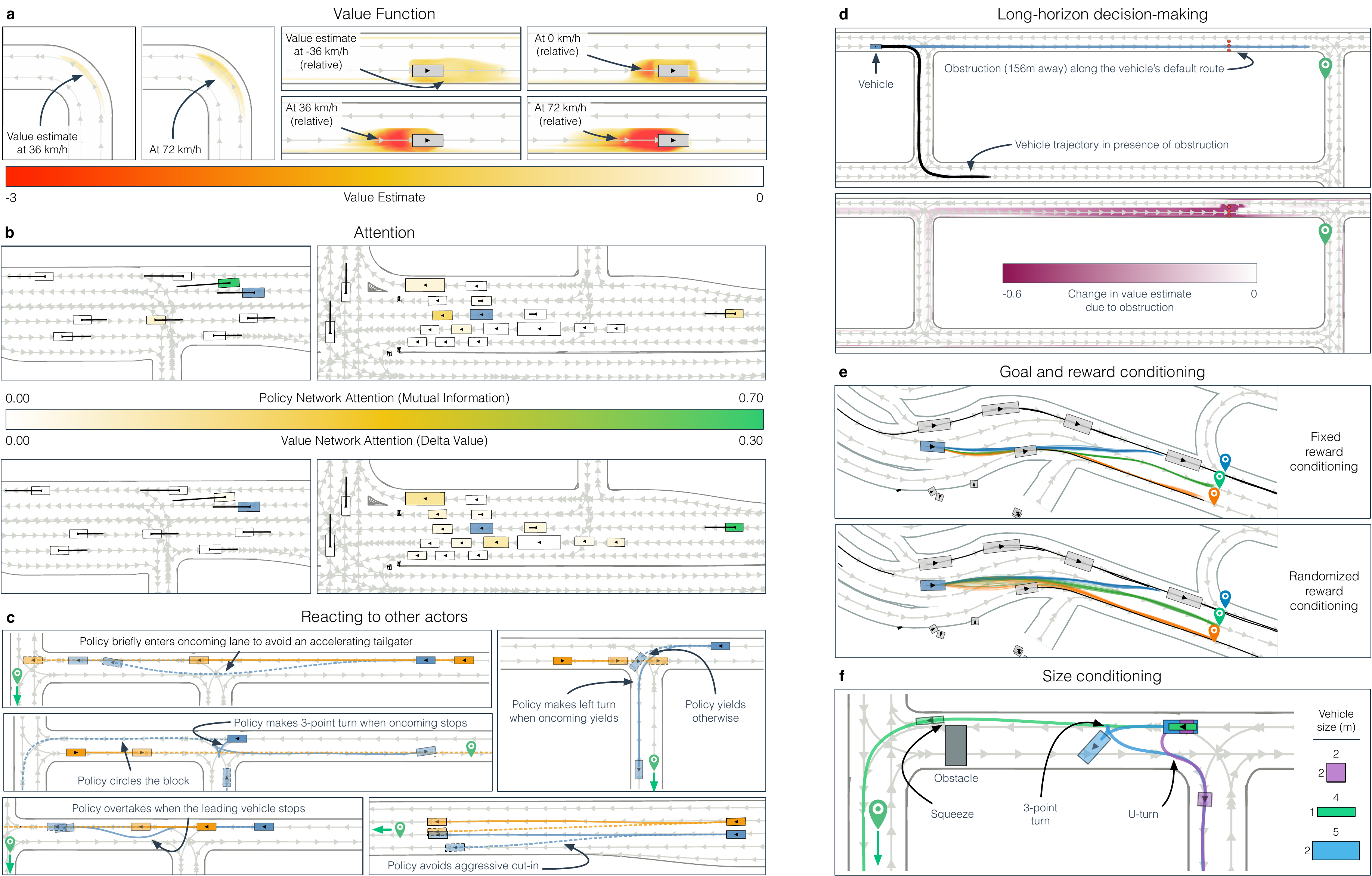}
\caption{
\textbf{Analysis of the learned policy and value networks.}
\textbf{a},
Value network estimates danger at fine granularity over the surface of the road. Heatmaps show state-value estimates for a driver placed at every possible location, oriented in parallel to the nearest lane.
Left: The network assigns lower value to rounding a sharp corner at higher speed.
Right: Value estimates for driving at various speeds in the vicinity of a vehicle (gray) moving 36 km/h to the right.
Being placed at rest (0~km/h) in front of the moving vehicle is likely to result in collision.
As velocity increases, the danger zone shifts behind the other vehicle. The estimated danger becomes acute as the relative speed increases.
\textbf{b},~Attention of the policy and value networks of the controlled vehicle (blue) to surrounding actors in two scenarios from the Waymo Open Motion Dataset.
Top row: Mutual information between the policy network's action distribution and actor presence.
Bottom row: Decrease in the value network's state-value estimate due to actor presence.
\textbf{c},
Scenarios where a scripted nearby vehicle (orange) either continues moving forward at constant velocity (solid orange line) or executes a different behavior (dashed orange line).
The policy (blue vehicle) executes maneuvers (solid blue line, dashed blue line) contingent on the other vehicle's behavior.
Final poses are translucent.
\textbf{d},
Policy and value networks perform long-horizon prediction and decision-making.
Top: Policy drives to its goal (20 second trajectory in blue) when the road is clear.
When an obstruction (black) is placed $156$~m away, the policy re-routes (black trajectory).
Bottom: Heatmap shows change in value estimate over the surface of the road due to introduction of the obstruction.
\textbf{e},
Top: Diversity of trajectories due to goal conditioning (three goals shown with associated trajectories in matching color).
Bottom: Additional variability due to randomized reward conditioning parameters.
\textbf{f},
Changing vehicle dimensions with all else fixed yields different behaviors:
A slender vehicle can squeeze around an obstacle; a compact vehicle with a tight turning radius can make a u-turn; a larger vehicle resorts to a three-point turn.
}
    \label{fig:route-analysis}
    \vspace{-2mm}
\end{figure*}

\ours training employs two neural networks:
the policy (actor) that chooses actions and the value function approximator (critic) that estimates the expected cost-to-go from a given state.
We examine how the policy's driving behavior changes over the course of training (\cref{fig:training-analysis}) and how the policy and value networks respond to targeted changes to their inputs in various scenarios (\cref{fig:route-analysis}).

\mypara{Reinforcement learning at scale yields mastery of complex skills.}
The scale of \ours training enables the policy to handle complex scenarios despite never seeing real-world or hand-designed driving scenarios during training.
The policy learns to execute unprotected left turns, drive in crowded roads used by both pedestrians and vehicles, and handle vehicles dangerously merging into the driver's lane (\cref{fig:training-analysis}a). In diagnostic tests designed for analysis, \ours vehicles are able to safely negotiate through a narrow bottleneck into a single lane when the other lanes are blocked by an accident (\cref{fig:training-analysis}b) and quickly merge three traffic flows into a single one due to road closures (\cref{fig:training-analysis}c).
Many of these skills are mastered only after $10^{11}$ to $10^{12}$ steps of training experience ($90$ to $1600$ million km driven).

\mypara{Value network detects dangerous states.}
To examine the value network's ability to detect dangerous states, we evaluate it on a
set of observations generated by densely sampling all possible positions and orientations of the driver on a fixed region of the map.
We find that the network appropriately assigns low value to states where the driver is taking a corner too fast, and where collision with another vehicle is imminent due to high relative velocity
(\cref{fig:route-analysis}a).

\mypara{Policy and value networks attend to salient scene features.}
Driving requires attending to the most consequential traffic participants at any given time, among hundreds of actors who may be present in the environment.
We assess the attention of our policy and value networks by analyzing the change in action distribution (via mutual information) and change in value estimate when each actor is individually removed (\cref{fig:route-analysis}b).
As expected, the networks sometimes attend to different actors:
For example, the value estimate is affected by all actors that make the scene more dangerous over the long term (for example a speeding car approaching a line of vehicles queued at a red traffic light), whereas the policy's action might not change due to such an actor if there is no way for the policy to mitigate the danger (\cref{fig:route-analysis}b).
\mypara{Policy executes maneuvers contingent on nearby traffic behavior.}
We evaluate the policy in scenarios where a nearby vehicle either behaves predictably (continuing to move at constant velocity) or unpredictably, with all else fixed.
We find that the policy executes appropriate discrete maneuvers contingent on the nearby vehicle behavior, like
changing lanes to avoid collision with a vehicle that is cutting into its lane and
passing a vehicle that unexpectedly stops in the road.
The policy executes contingent longer-term routing maneuvers, like turning around by circling the block instead of making a three-point turn, depending on traffic (\cref{fig:route-analysis}c).

\mypara{Policy reacts to potential events far in the future.}
Both networks are also able to react to salient scene features even when they are distant, like an obstruction $150$\,\si{\meter} down the road (\cref{fig:route-analysis}d),
and can ignore actors that are near but irrelevant (\eg a car parked few meters from the driver in a parking lot).
While considering potential events beyond the planning horizon is often a challenge for trajectory-based planners~\cite{casas2021mp3}, the \ours policy optimizes long-term return directly without the limitations of a short time horizon.

\mypara{One policy learns a continuum of driving styles.}
Reward function coefficients $C_\text{reward}$, vehicle dimensions, and vehicle goals are all randomized during training.
As a result, the policy learns a parameterized family of driving styles (\cref{fig:route-analysis}e,f; \cref{fig:conditional-analysis}); different styles can be elicited from a single trained policy by setting the parameters accordingly, without any retraining or fine-tuning. %
For example, the policy squeezes through narrow passages or performs tight turns if and only if the vehicle dimensions permit this (\cref{fig:route-analysis}f).
Likewise, reducing the conditioning parameter that controls sensitivity to red lights makes the policy more willing to run red lights in order to accomplish other goals.

\mypara{\ours yields a highly efficient, capable, and realistic simulation environment.}
The \ours training configuration features substantial dynamics noise and diverse reward conditioning parameters $C_\text{reward}$ (see \cref{sec:technical}).
We can configure the same simulation infrastructure for long-form evaluation of trained policies. In this configuration, we reduce the injected dynamics noise, increase control frequency, and set conditioning parameters $C_\text{reward}$ that prioritize safety for all actors. This yields a fast, cost-effective, and highly robust traffic simulator.
In this regime, a fully trained \ours agent experiences on average 17.5 years of driving and travels over 3 million \si{\kilo\meter} before encountering an incident.
(For reference, human drivers average approximately $\num{829000}$ vehicle kilometers traveled per police-reported traffic crash in the United States~\cite{Stewart2023}, or as much as 1 crash per $\num{24800}$ \si{\kilo\meter} in narrower domains such as San Francisco~\cite{Flannagan2023}.)
To our knowledge this is the first demonstration of long-term robust traffic simulation based on independent agents traversing a diverse urban road network.

We evaluate the realism of the driving behaviors learned via \ours on the Waymo Open Sim Agents Challenge (WOSAC),
which measures the ability to reproduce real-world driving behaviors for simulation purposes~\cite{montali2023waymo}.
Despite not using any human data for training, the \ours policy exhibits many characteristics of human driving, achieving a score of 0.62 in zero-shot evaluation on the realism meta-metric, outperforming several approaches based on supervised autoregressive prediction. (Details in \cref{sec:simagents}.)

\section{Discussion}\label{conclusion}

Many questions remain to fully understand the long-term role of self-play in delivering broad-competence robust autonomy.
First, our work has been conducted entirely in simulation.
Techniques for transferring policies from simulation to reality will have to be brought to bear before claims can be made regarding the efficacy of self-play policies in the physical world~\cite{Mueller2018:CoRL,Lee2020:SciRob,Kaufmann2023}.

Second, our work has focused on planning and decision-making, largely abstracting the perception stack.
To integrate the presented findings into an operational system, sensing and perception will have to be modeled much more closely.
An exciting possibility is to combine large-scale self-play training with data-driven simulation of the associated perceptual inputs (\eg, camera images)~\cite{Ost2021:CVPR,Yang2023:CVPR,hu2023gaia1}.
It is likely feasible in the coming years due to ongoing improvements in simulation methodology~\cite{Shacklett2021,petrenko2021megaverse,shacklett23madrona}, computing hardware, and system architectures.
Combining self-play with photorealistic sensor simulation would substantially increase the computational footprint of each experience, but the wall-clock training time can be maintained by scaling out over a commensurate number of compute nodes~\cite{llama3}.

Third, our work has demonstrated that training without real-world driving traces can yield policies that are surprisingly human-like~\cite{montali2023waymo} and highly robust when tested in recorded real-world scenarios with human participants~\cite{caesar2022nuplan,gulino2023waymax}.
By contrast, common perspectives on learning-based autonomous driving hold that recorded datasets will play a key role in training driving policies~\cite{Jain2021,Hawke2021,chen2023e2esurvey}.
How do we reconcile our findings with these views? One possibility is to combine large-scale self-play training with training on recorded scenarios, perhaps via a combination of reinforcement learning and imitation learning~\cite{Lu2023,Zhang2023:CoRL}.
This can further increase robustness and help bridge simulation and reality.

Our findings may inspire broader application of self-play in training agents that act in the presence of (and in close coordination with) humans in physical and digital environments. Such coordinated action may be called for in mobile robotics, in both consumer and industrial settings, and in digital domains such as online games. We have shown that policies that function effectively in the presence of human actors in complex dynamic environments can be trained without utilizing human data. Broader application of this methodology may substantially reduce the cost and complexity of training autonomous policies by meaningfully reducing the need for human data collection.

\bibliography{bibliography}
\bibliographystyle{icml2024}

\clearpage

\begin{appendices}
\crefalias{section}{appendix}

\section{Simulator Design}
\label{sec:sim_design}

\ours is a \textit{batched} simulator~\cite{isaacgym,freeman2021brax,petrenko2021megaverse,Shacklett2021}, implemented in PyTorch~\cite{Ansel_PyTorch_2_Faster_2024} and designed for GPU acceleration. A single instance of \ours simulates thousands of worlds, enabling it to leverage the parallelism of modern GPUs. Implementing \ours efficiently required the development of custom batched operators for the kinematics, collision checking, initialization of urban driving environments, and many other features. Below we outline the overall flow of \ours and describe the acceleration techniques used.

At every timestep, \ours takes the current state of the $i$-th world $s_i^{(t)}$ and vehicle controls $a_i^{(t)}$ to produce the state at the next timestep $s_{i}^{(t+1)}$. Instead of running multiple copies of the environment to simulate $N=\num{38400}$ worlds, a single instance of \ours simulates $N$ worlds in parallel. The state and vehicle controls of \emph{all} $N$ worlds are denoted $\mathbf{s}^{(t)}$ and $\mathbf{a}^{(t)}$.% to refer to the state and vehicle controls, respectively, of \emph{all} worlds.

\subsection{World initialization}

The simulation process begins with the initialization of urban driving environments by placing up to $N_a = 150$ vehicles at random positions on the map.
We first draw a random sample over map locations, vehicle headings, and vehicle bounding box dimensions and reject states that are off-road (off-road checking is detailed below). A naive application of this process results in a marginal distribution of vehicle locations that is biased towards wider road sections. To correct for this, we estimate this marginal distribution from an initial sample and use it to adjust the proposal distribution used in subsequent rejection sampling.
Given the set of valid vehicle states, we then select a collision-free subset with the desired number of agents (collision detection is detailed below). This subset becomes the initial traffic configuration at $t=0$.

For each retained vehicle, we select a sequence of its waypoints (goals). The first waypoint is sampled uniformly over the map and additional waypoints are sampled such that given the $j$th waypoint, the $(j+1)$th waypoint is at least \SI{20}{\meter} away and no more than \SI{200}{\meter} away, and has lane heading that is within 60 degrees of the $j$th waypoint's lane heading. There are cases where the $j$th waypoint is in a location where these constraints cannot be met (\eg a dead end). In these cases, we gradually relax the constraints as we try to sample points that fit them. The intent of this sampling procedure is to generate waypoint sequences that resemble realistic driving routes where intermediate destinations are reached in a natural succession.

We use two acceleration techniques specific to initialization. First, we draw a large buffer of vehicle states and goals all at once and then consume that buffer until it is empty as new initializations are requested, \eg when episodes reset. This allows for better cost amortization when generating initial states.
Second, we use sequential rejection sampling to produce collision-free initializations. We add one agent to the world, then add a second agent that is not in collision with existing agents (via rejection sampling), and so on until we reach the desired number of agents. This approach was necessary because the probability that an independent sample of over one hundred vehicles contains no collisions is extremely low for small maps.

\subsection{Dynamics model update}

The dynamics model (described below)
produces $\mathbf{s}^{(t+1)}$ given $\mathbf{s}^{(t)}$ and $\mathbf{a}^{(t)}$. This is a set of element wise operations (trigonometric functions, multiplications, divisions, \textit{etc.}) that are parallelized on a GPU using their respective PyTorch implementations.

\subsection{Road localization}

The next step is to localize $\mathbf{s}^{(t+1)}$ on the road surface.

\mypara{Road representation.}
\ours represents the road surface as a set of potentially overlapping polygons, exclusively using convex quadrilaterals. We find quadrilaterals to offer a good balance between expressivity and the simplicity of operations. A given lane on the road surface is
approximated by quadrilaterals that have the same width as the lane and are \SI{1}{\meter} in length.
We find this resolution of polygons to be a good trade-off between the accuracy of road geometry approximation and the total number of primitives.

It should be noted that the number of geometric primitives could be minimized further by merging polygons in the regions with simple geometry, such as straight road segments. However, we retain the uniform polygon density irrespective of the geometric complexity because this polygonal subdivision additionally serves as a spatial hash map for certain types of queries (\eg map observations are pre-computed for all polygon center points).

\mypara{Frenet coordinates.}
Let $q$ represent the distance along a lane, $d$ be the distance from lane center, and $\texttt{polyId}$ be a unique identifier of the polygon approximating the road geometry at the current location. The Frenet coordinate for position $(x, y)$ is then $(q, d, \texttt{polyId})$.

We convert world-frame state $\mathbf{s}^{(t)}$ to Frenet-frame state $\mathbf{f}^{(t)}$ as this information is useful for constructing actor observations and covers off-road checking in the majority of cases. We construct the Frenet-frame state by first finding the polygon that contains $(x, y)$. Given this polygon, we can compute $(q, d)$ by transforming $(x, y)$ into the coordinate frame defined by the polygon's heading and center point then adding the distance from the start of the lane to the polygon's center to $q$.
If multiple polygons contain $(x, y)$, we use the polygon where the agent is most lane centered (where $d$ is the smallest).

Naively implementing this would require performing $O(N \times A \times P)$  point-in-polygon checks, where $N$ is the number of worlds, $A$ is the number of agents, and $P$ is the number of polygons. This is prohibitively computationally expensive.

\mypara{Spatial hashing.}
We accelerate this via spatial hashing on the GPU. Specifically, we first construct a 2-D grid of non-overlapping axis-aligned boxes with a fixed width and height. This scheme allows us to trivially map a point $(x, y)$ to its given bucket in the hash map. We then assign each polygon in the road representation to the (potentially multiple) buckets that it overlaps. Note that this step only needs to be performed once at startup.

We localize $(x, y)$ by first retrieving the polygons that are in its bucket and then performing a point-in-polygon check for each matching quadrilateral. One challenge of implementing this efficiently is that each hash bucket has a different number of corresponding polygons. We maintain efficiency by designing APIs to expect these ``ragged'' variable-size tensors and only perform padding on operations that require it (\eg finding the polygon that minimizes $d$).

This spatial hash additionally allows for the support of multiple maps via augmentation of vehicle coordinates $(x, y)$ with the map Id: $(x, y, \texttt{mapId})$. Essentially, the map an agent is driving on functions as a ``third'' dimension, allowing us to filter out polygons from other maps at the hash bucket retrieval step.

\subsection{Off-road checking}
\label{methods:off-road}

Given $\mathbf{s}^{(t)}$, we localize the vehicle's bounding box center and corners on the road surface using the spatial hash procedure described above. In the majority of cases, a vehicle is off-road if any of these 5 points could not be localized onto the road surface. However, there are two edge cases that are important to handle.

\mypara{Curved roads and islands.}
These are various situations where all 5 of these points can be on the road surface but the vehicle should still be considered off-road. Two such examples are curved roads (where part of the bounding box overhangs the edge of the road) and pedestrian safety islands (where the vehicle can straddle the islands). To handle both these cases, we generate a set of out-of-bounds (OOB) points by taking the mid-point of each polygon edge, nudging it slightly in the outwards direction, and keeping all points that do not lie within any other polygon. A vehicle is classified as off-road if any of these OOB points are found to be within its bounding box.

We again use our spatial hashing data structure to accelerate the check between vehicle bounding box and the OOB points. Specifically, each OOB point is first associated with its hash bucket. Then, we lookup which buckets the vehicle overlaps with and perform a point-in-bounding-box check with that set of points.

We associate the vehicle bounding box with appropriate hash buckets as follows: first, we set the bucket size to be $2\times$ the maximum possible vehicle length. Then, we approximate the vehicle's oriented bounding box by its axis-aligned bounding box (AABB). Due to our choice of spatial hash bucket size, the AABB overlaps with at most 4 buckets, each containing (at least) one of the AABB's corners. We retrieve the OOB points that correspond to $\le 4$ buckets overlapped by the vehicle and perform a point-in-bounding-box check with each. We find that this fairly simple approximation is appropriate since it enables very fast lookup which is crucial as it is performed at every simulation step.

\mypara{Gaps in the road surface.}
Due to the extremely low off-road rates achieved by \ours policies, we found that most remaining off-road events are characterized by thin gaps in the road surface (sometimes under $\SI{1}{\milli\meter}$), often located between neighboring lanes. These thin gaps are not actual geometrical features but numerical inaccuracies in the underlying map files.

During training, we allow any of the 5 points used for off-road checking to be in one of the spurious gaps by measuring the distance from the point to the closest road polygon and considering points within $\delta=\SI{15}{\centi\meter}$ to be still on the road surface.
We again use spatial hashing to accelerate this lookup. We take edges of all road polygons and assign them to any bucket that they overlap with or pass within the distance $\delta$ of, so we could quickly obtain a list of candidate edges.

\subsection{Collision detection}
\label{methods:collision}

Given $\mathbf{s}^{(t)}$ and $\mathbf{s}^{(t+1)}$, we detect collisions as follows. Let $s_{a_i}^{(t)}$ and
$s_{a_i}^{(t+1)}$ be successive states of agent $i$, and  $s_{a_j}^{(t)}$ and
$s_{a_j}^{(t+1)}$ be the successive states of agent $j$. We perform two checks to see if a collision occurred. First we transform $s_{a_j}^{(t)}$ into the coordinate frame of $s_{a_i}^{(t)}$ and $s_{a_j}^{(t+1)}$ into the coordinate frame of $s_{a_i}^{(t+1)}$. Then we check to see if any of the lines defined by the movement of agent $j$'s corners intersectv with agent $i$'s bounding box (the bounding box is centered at the origin). We then swap the roles of agents $i$ and $j$ and perform this check again. A collision occurred between agents $i$ and $j$ if either check is positive.
Note that we only perform collision detection, not collision simulation.

We accelerate this using our spatial hash by constructing, for all agents, the axis-aligned bounding box (AABB) that contains the vehicle's bounding box at \emph{both} states $s^{(t)}$ and $s^{(t+1)}$. We assign the vehicle to all buckets that overlap with this bounding box and perform the collision detection procedure described above for all pairs of vehicles that share at least one bucket. We use the world ID of each agent as an additional dimension in the spatial hash so that vehicles can only collide with other vehicles in the same world. This hashing method is fast to compute and greatly reduces the number of candidate collision pairs. Therefore hashing represents a large improvement over a naive pairwise check.

The check described above only works when (at least) one vehicle is in motion. This is acceptable during simulation, as at least some movement is required to transition from collision-free to a colliding state. For detecting collisions at $t=0$, we simply check to see if the two bounding boxes overlap at state $s^{(0)}$. We use the spatial hash in a similar manner as before, except the AABB corresponds to a single (initial) state.

\subsection{2.5-D simulation}

Simulating driving can be largely approximated without errors as a two-dimensional problem. This approximation enables performance improvements and reduces code complexity (thereby limiting the surface area for coding errors). However, certain cases cannot be accurately approximated in two dimensions, like overpasses. \ours handles this with ``2.5-D'' simulation. We simulate the world as if it were 2-D and then correct for these errors. For example, we perform collision detection in two dimensions and then remove collisions that would not have occurred in 3-D.  We maintain the vehicle's $z$-coordinate by applying the dynamics model purely in 2-D and then looking up new $\mathbf{z}^{(t+1)}$ for all vehicles from the map that correspond to new locations $\mathbf{s}^{(t+1)}$. We use $\mathbf{z}^{(t)}$, $\mathbf{s}^{(t)}$, and the maximum slope of a given map to filter out conflicting $z$ values in cases like overpasses.

\subsection{Hardening}

Given the extremely low collision and off-road rates seen in \ours training we found that extremely rare bugs would dominate the collisions and off-road events when present. We iterated extensively; training numerous policies to very high fidelity, watching videos of collisions and off-road events, and sieving those caused by coding errors and numerical inaccuracies rather than the agent's poor decision making. Each iteration would yield new bug fixes, improving both training and downstream benchmark performance.

To address these bugs, we built numerous visualization tools, as merely knowing that a rare bug existed was not enough to fix it — we had to find the exact steps to reproduce it as well. The visualization, diagnostic, and recording tools we developed were instrumental to the overall success of the project.

We additionally developed multiple simplified versions of \ours to speed up diagnostics and troubleshooting. Using the minimal single-agent configuration we could train policies to convergence in under 20 minutes, allowing us to get immediate feedback on the core functionality.
We established infrastructure for continuous testing, repeatedly retraining policies from scratch using these streamlined simulator configurations.
This system has significantly accelerated the identification and isolation of regressions, thereby conserving numerous development cycles.

\section{Defining the partially observed stochastic game}
\label{sec:pos_game}

We model the environment that our agents learn in as a \emph{partially observed stochastic game} (POSG)~\cite{hansen2004dynamic}; an extension of POMDPs to the multi-agent setting in which there are multiple agents with conflicting goals. We define each of the components of the POSG below.

\subsection{Observations}
\label{methods:obs-acts-rew}

We render the world state $s^{(t)}$ into a relatively low dimensional vector representation. In order to drive safely, a \ours agent utilizes information about vehicle's dynamics and its position w.r.t. the lane $S^{(t)}$, approximate map of the surrounding area, including roads and traffic lights $\left( W^{(t)}_{\text{lane}}, W^{(t)}_{\text{boundary}}, W^{(t)}_{\text{stop}} \right)$, observations of other traffic participants $A^{(t)}$, desired destination and intermediate waypoints $G^{(t)}$, as well as the agent's conditioning $C_\text{reward}$ (\cref{tab:reward_rand}) .

Local observation $S^{(t)}$ can be further broken down as follows (time indices omitted for clarity):

\begin{itemize}
    \item $c, \, \theta$: the distance from the current lane center and the angle relative to lane heading.
    \item $\kappa$: local road curvature.
    \item $v$: current speed of the vehicle.
    \item $v_\text{lim}$: maximum allowed speed.
    \item $\phi$: current steering angle.
    \item $a_\text{long}, a_\text{lat}$: current longitudinal and lateral acceleration.
    \item Driver's acceleration limits $C_\text{acc}$.
    \item $C_\text{throttle}, \, C_\text{steer}$: randomized coefficients determining the vehicle's responsiveness to throttle and steering inputs.
    \item $l, \, w$: driver's vehicle's length and width.

\end{itemize}
Note that all of these observations are normalized to $[-1,1]$ range and provided in egocentric frame.

Map observations outline the surrounding area at two levels of resolution, including the high-level overview of the nearby road network $W^{(t)}_{\text{lane}}$ and a more detailed set of features $W^{(t)}_{\text{boundary}}$ that represent the precise shape of the local driveable area.

The coarse map view $W^{(t)}_{\text{lane}}$ provides a set of positional features sampled along driveable lanes at \SI{40}{\meter} intervals, containing information about lane widths and heading directions. These features also convey higher-level routing information similar to that provided by GPS-based navigation applications. Each observation within $W^{(t)}_{\text{lane}}$ is augmented with relative (w.r.t. other observations) and absolute normalized distances to the next goal, which allows \ours agents to make informed routing decisions at complex junctions. We pre-calculate all routing information by running the Dijsktra's algorithm~\cite{dijkstra1959note} offline, computing pairwise distances between all locations of interest.

For a more fine-grained road geometry representation $W^{(t)}_{\text{boundary}}$ we provide midpoints of nearby polygon edges  spaced roughly \SI{1}{\meter} apart which delineate the out-of-bounds areas closest to the driver (Fig. \ref{fig:technical}c). For both $W^{(t)}_{\text{lane}}$ and $W^{(t)}_{\text{boundary}}$ we provide the $80$ features closest to the driver, additionally limited to \SI{200}{\meter} viewing horizon for the coarse map.
We pre-calculate $W_{\text{lane}}$ and $W_{\text{boundary}}$ for each map polygon which enables extremely fast retrieval at runtime, only requiring a simple conversion to driver's egocentric frame to obtain $W^{(t)}_{\text{lane}}$ and $W^{(t)}_{\text{boundary}}$.

For observations of other agents we return the $N_o$ nearest agents within $\delta_\text{max}=\SI{200}{\meter}$ of the driver. We pad the array of observations in case there are fewer than $N_o$ nearby actors. For each agent, we observe its position, orientation, velocity, dimensions (i.e. width and length), and z-coordinate, all transformed into driver's egocentric frame. The goals, momentary accelerations, dynamics properties, or conditioning parameters of other agents are not observed.

While we do not explicitly model occlusions, we keep the maximum number of observed agents during training relatively low: $N_o=20 \ll N_a$. Together with the distance-based visibility threshold $\delta_\text{max}$ this allows us to train the agent well adjusted to limited observability. Our permutation-invariant model architecture allows the value $N_o$ to be trivially increased at test time to use all available information (see the model architecture details below).

\subsection{Actions and dynamics}

Our agents use a discrete set of actions to control the vehicle's change in acceleration using a jerk-actuated bicycle dynamics model. The action space includes $12$ total actions, the Cartesian product of the sets of available values of longitudinal $\dot{a}_\text{long} \in \{-15, -4, 0, 4\}$ \si{\meter\per\second\cubed} and lateral jerk $\dot{a}_\text{lat} \in \{-4, 0, 4\}$ \si{\meter\per\second\cubed}.

We compute longitudinal and lateral accelerations using numerical integration ($\Delta t = \SI{0.3}{\second}$ during training):

\begin{align}
    a_{\text{long}}^{(t)} &= a_{\text{long}}^{(t-1)} + C_\text{throttle} \, \dot{a}_\text{long} \, \Delta t
    \label{eq:long_accel}
    \\
    a_{\text{lat}}^{(t)} &=  a_{\text{lat}}^{(t-1)} + C_\text{steer} \,  \dot{a}_\text{lat} \, \Delta t
    \label{eq:lat_accel}
\end{align}
where coefficients $C_\text{throttle}, C_\text{steer} \in C_{\text{dynamics}}$ are sampled from a mixed uniform distribution $X(1.25)$, defined as:
\begin{align*}
X(a) &= 0.5\,U( a^{-1},1 ) + 0.5\,U(1, a), \quad a > 1
\end{align*}
This distribution generates an equal number of samples smaller and greater than one, thereby allowing for a balanced randomization of dynamics properties.

We apply a small modification to the \cref{eq:long_accel,eq:lat_accel}, setting values $a_{\text{long}}^{(t)}$, $a_{\text{lat}}^{(t)}$ to exactly $0$ when acceleration changes sign (i.e. when $a^{(t-1)}_{\text{long}} \, a^{(t)}_{\text{long}} < 0$). We found that this modification makes it easier for the agent to wait in place or drive at a constant velocity, producing smoother trajectories.

The acceleration components are then clipped ensuring the g-forces stay within the specified limits (here $C_{\text{acc}}\sim~X(1.5)$):
\begin{align*}
    a^{(t)}_{\text{long}} &\leftarrow \texttt{clip} \left( a^{(t)}_{\text{long}},\, -5,\, 2.5\,C_{\text{acc}} \right) \\
    a^{(t)}_{\text{lat}} &\leftarrow \texttt{clip} \left( a^{(t)}_{\text{lat}},\, -4,\, 4 \right)
\end{align*}

We update the velocity magnitude using the trapezoidal rule (averaging previous and current accelerations):
\begin{align*}
    v^{(t)} &= v^{(t-1)} + 0.5 \, \left( a_{\text{long}}^{(t)} + a_{\text{long}}^{(t-1)} \right) \Delta t
\end{align*}
Just as for the accelerations, we set $v^{(t)}$ to exactly $0$ when its value changes sign.
We then clip $v^{(t)}$ to stay within the randomized speed limit (${C_{\text{vel}} \sim~X(1.5)}$):
\begin{align*}
    v^{(t)} &\leftarrow \texttt{clip}\left( v^{(t)},\, -2,\, 20\,C_{\text{vel}} \right)
\end{align*}

To reach the acceleration $a^{(t)}_{\text{lat}}$, the vehicle would have to follow the arc with radius $|\rho|$ and signed curvature $\rho^{-1}$ applying the steering angle $\phi$ (here $l_\text{wb}$ is the vehicle's wheelbase and $\epsilon=10^{-5}$ ensures numerical stability):
\begin{align*}
    \rho^{-1} &= \frac{a_{\text{lat}}}{\max(v^2, \epsilon)}
    \\
    \rho^{-1} &\leftarrow  \text{sign}\left( \rho^{-1} \right)\, \max\left( \left| \rho^{-1} \right|, \epsilon \right) \\
    \phi &= \arctan \left( \rho^{-1} \, l_{\text{wb}} \right)
\end{align*}

We calculate the change in the steering angle, $\delta_\phi$, and update the effective steering angle, $\phi^{(t)}$, ensuring that both values remain within conservatively defined limits ($\delta_{\text{max}}=\SI{0.6}{\radian\per\second}, \phi_{\text{max}}=\SI{0.55}{\radian}$):
\begin{align}
    \delta_\phi &= \texttt{clip}(\phi - \phi^{(t-1)},\, -\delta_{\text{max}}\,\Delta t,\, \delta_{\text{max}}\,\Delta t)
    \label{eq:delta-phi}
    \\
    \phi^{(t)} &= \texttt{clip}(\phi^{(t-1)} + \delta_\phi,\, -\phi_{\text{max}},\,\phi_{\text{max}})
    \label{eq:new-phi}
\end{align}
The clipping in \cref{eq:delta-phi,eq:new-phi} prevent excessive changes in the steering angle, limiting the maximum lateral acceleration at low speed. To account for the limited steering actuation, we update the effective signed curvature $\rho^{-1}$ and  acceleration $a_{\text{lat}}^{(t)}$ accordingly:
\begin{align*}
    \rho^{-1} &\leftarrow \frac{\tan\left(\phi^{(t)} \right)}{l_{\text{wb}}}
    \\
    a_{\text{lat}}^{(t)} &\leftarrow \left(v^{(t)}\right)^2 \rho^{-1}
\end{align*}

Finally, the resulting movement of the vehicle is updated using the bicycle dynamics model:
\begin{align*}
    d &= 0.5 \, \left( v^{(t)} + v^{(t-1)}\right) \Delta t \\
    \theta &= d \, \rho^{-1}
    \\
    \Delta x &= \rho \sin \left(\theta \right) \\
    \Delta y &= \rho \cos \left(\theta \right)
\end{align*}
where $d$ is the displacement along the arc and $\theta$ is the angular displacement.

\subsection{Reward}

Our agents optimize a scalar reward function consisting of multiple terms weighted by their respective coefficients $\alpha_{\left(\cdot\right)}$ which are randomly sampled at the beginning of each episode (see \cref{tab:reward_rand}). This reward function $R$ is defined by:
\begin{equation*}
\begin{aligned}
    R &=R_\text{goal} + R_\text{collision}+ R_\text{off-road} + R_\text{comfort} + R_\text{lane} \\
      &+ R_\text{velocity} + R_\text{reverse} + R_\text{stop-line} + R_\text{timestep}
\end{aligned}
\end{equation*}
where the individual terms can be described as follows:
\begin{itemize}
    \item $R_\text{goal}$ rewards the agent for reaching its intermediate goals (waypoints) and the final goal.
    \item $R_\text{collision}$ penalizes agents for colliding with vehicles or pedestrians with an additional penalty for colliding at high speed. Randomizing $\alpha_{\text{collision}}$ allows us to populate the training environments with agents of varying risk tolerance, from very aggressive to very conservative.
    \item $R_\text{off-road}$ penalizes agents for leaving the road.
    \item $R_\text{comfort}$ is a penalty for exceeding the comfortable limits of acceleration and jerk. Randomized $\alpha_{\text{comfort}}$ creates a distribution of driving styles, from relatively smooth to practically unconcerned with comfort.
    \item $R_\text{l-align}$ incorporates the preferences for driving in the designated driving direction and staying \textbf{parallel} to road lanes. We randomize this term in a wide range, which occasionally exposes our agent to erratic actors driving against traffic.
    \item $R_\text{l-center}$ rewards the agent for staying \textbf{centered} in the lane. We randomize $\alpha_{\text{center-bias}}$ for additional behavioral diversity.
    \item $R_\text{velocity}$ encourages forward progress and motivates the agent to prefer routes with consistent traffic flow over traffic jams. We found that this term accelerates convergence in the early stages of training and helps us train policies that are better at avoiding gridlocks in self-play.
    \item $R_\text{reverse}$ penalizes the agents for reversing. Due to the randomization of $\alpha_{\text{reverse}}$ some agents perform multi-point turns to reach the goals behind them, while others prefer to drive forward and wrap around the block or perform a U-turn instead.
    \item $R_\text{stop-line}$ penalizes the agent for crossing a stop line at a red light.
    \item $R_\text{timestep}$ is a small penalty applied at each simulation step. This serves a function similar to the discount $\gamma$, except we disable $R_\text{timestep}$ when ego is stationary which creates agents more willing to patiently wait at traffic lights and intersections when necessary.
\end{itemize}

\subsection{Additional randomized components}

In addition to the reward function coefficients we randomize a number of simulator's components to generate diverse embodiments and behavioral patterns. Agents have access to the randomized parameters as a part of their conditioning.

\mypara{Vehicle size.}
We randomize vehicle's dimensions as follows (measured in meters):
\begin{itemize}
    \item    Vehicle length and wheelbase: $l \sim U(0.8, 7)$, $l_{\text{wb}} = 0.6\,l$.
    \item    Vehicle width: $w \sim U(0.8, 3)$.
    \item    The vehicle width is clipped: $w \leftarrow \text{min}(w, l)$.
\end{itemize}

Size randomization allows us to generate a variety of road users, from very maneuverable agents with a footprint of a pedestrian to relatively big vehicles. We represent occasional bigger vehicles at test time using a set of smaller bounding boxes (\eg a semi-truck with a trailer can be split into two bounding boxes).

\mypara{Goals.}
A final goal is sampled for each agent. Additionally we sample $N_{\text{wp}} \sim U\{0,3\}$ intermediate waypoints (see ``World Initialization'' for details).
We randomize the goal collection radius $\delta_{\text{goal}}$ to control how precisely our driver is required to adhere to the designated route (see \cref{tab:reward_rand}).

\mypara{Vehicle dynamics.}
For each agent in each episode we randomly sample maximum permitted acceleration, velocity, throttle and steering coefficients (refer to ``Actions and dynamics'' for specific randomization ranges).

\mypara{Traffic lights.}
Real driving situations include a diverse set of traffic light systems, from synchronized lights at 3-way and 4-way intersections to occasional faulty, disabled, or miscalibrated signals.
Instead of explicitly modeling this distribution in \ours we simply pick random durations of green, yellow, and red signals for each traffic light independently. We start with default signal durations used in CARLA (typically $\hat{\tau}_{\text{red}}=2$, $\hat{\tau}_{\text{yellow}}=3$, and $\hat{\tau}_{\text{green}}=10$ seconds) and then randomize them for each episode within the following ranges:

\begin{itemize}
    \item $\tau_{\text{red}} \sim U(0.15\,\hat{\tau}_{\text{red}}, 5.0\,\hat{\tau}_{\text{red}})$
    \item $\tau_{\text{yellow}} \sim U(0.5\,\hat{\tau}_{\text{yellow}}, 0.75\,\hat{\tau}_{\text{yellow}})$
    \item $\tau_{\text{green}} \sim U(0.1\,\hat{\tau}_{\text{green}}, \hat{\tau}_{\text{green}})$
\end{itemize}

We additionally remove 20\% of individual lights, 20\% of traffic light groups (\eg all lights at an intersection), and in 20\% of all episodes we disable traffic lights entirely, making all intersections unregulated. We also set 5\% of all remaining traffic lights to be constantly green.

Even though we train only on $128$ variants of CARLA maps ($8$ maps modified with affine transformations), this aggressive traffic light randomization allows us to combinatorially increase the total number of unique training environments and traffic patterns, preventing overfitting. Our policy learns to handle arbitrary traffic light configurations which helps significantly in benchmarks like nuPlan where traffic signal states are derived from the sense stack and often have errors.

The maximum red light duration in our simulator is deliberately limited ($\tau_{\text{red}} \le \SI{10}{\second}$). This allows us to mitigate the impatient nature of discounted reward maximization. When durations of required stops are short during training, agents with $\alpha_{\text{stop-line}} \gg 0$ are never incentivized to run red lights to maximize their return by getting to the goal faster. We find that our agents, due to their reactive nature, generalize to much longer red light durations occasionally encountered in benchmarks.

\mypara{Modeling erratic drivers.}
We aim to train an attentive, rational, and defensive driver which does not require cooperation of other traffic participants to ensure safety. Ideally, even clearly erratic maneuvers by other vehicles should not lead to collisions or comfort violations on behalf of \ours agent.
In order to achieve this, we degrade a fraction of agents during training by introducing two types of modifications:
\begin{enumerate}
    \item Up to 5\% of agents occasionally do not see other vehicles. This models inattentive drivers and drivers with blind spots.
    \item Up to 10\% of agents sharply apply their brakes at arbitrary moments for a short duration before resuming normal driving. This models drivers that stop without warning or remain stationary when a traffic light turns green.
\end{enumerate}
Between random applications of these modifications, agents are still controlled by \ours policy and are completely oblivious to their modification, which makes it impossible for regular drivers to predict their erratic behavior. We exclude trajectories from these agents from the rollouts used for training.

\section{Training algorithm}
\label{sec:training_algo}
Agents are trained using a version of Proximal Policy Optimization (PPO)~\cite{schulman2017ppo} derived from the Stable Baselines codebase~\cite{raffin2021stable}. PPO trains a policy that outputs actions, and a critic that estimates discounted returns. In our implementation, we do not share any parameters between the policy and the critic, which empirically produces comparatively more robust low entropy policies at convergence.
Additionally, we add the terminal value estimate to reward at the end of all truncated episodes, thus emulating infinite-horizon learning, similar to ``Reset Handling'' in \citet{rudin2021walkinminutes}. We use Adam optimizer~\cite{kingma2015adam} with annealed cosine-rate learning schedule:
\begin{equation*}
    \alpha^{(k)} = \frac{\alpha^{(0)}}{2} \, \left[1 - \cos\left( \pi - \frac{\pi\,k}{K}\right)\right]
\end{equation*}
where $k$ is the current training iteration and $K$ is the maximum number of iterations.
We use in DD-PPO~\cite{Wijmans2020} to train with mutliple GPUS. Specifically, each GPU collects experience independently and then gradients are synchronized before the model update.

During development, we optimized a subset of hyperparameters of PPO and \ours simulator using Population Based Training (PBT)~\cite{jaderberg2017population,petrenko2023dexpbt}. We conducted this optimization using a simplified version of \ours training setup (single map, reduced randomization) which enabled faster iteration. Our PBT experiments informed a number of non-trivial hyperparameter choices, from the learning rate schedule to comparatively large integration step $\Delta t = \SI{0.3}{\second}$.
\cref{tab:ppo_params} provides our final list of training hyperparameters.

\mypara{Advantage filtering.}
To make use of the vast scale of synthesized data, we implement a technique termed \textit{advantage filtering} which can be viewed as a variant of Prioritized Experience Replay~\cite{schaul2016prioritized} where the majority of transitions are sampled exactly \textbf{zero times}.

Specifically, for each transition we calculate the advantage estimate $\hat{A}^{(t)}_{\text{GAE}}$~\cite{schulman2016gae} and then simply discard all transitions that satisfy $\left|\hat{A}^{(t)}_{\text{GAE}}\right| < \eta$. The adaptive filtering threshold $\eta$ is set to 1\% of the moving average estimate of the maximum advantage magnitude at the current stage of training, thus rendering the method insensitive to the absolute scale of rewards (see \Cref{alg:advfilter}).

\algnewcommand\Var[1]{\mbox{\itshape#1}}
\algnewcommand\Varm[1]{\ensuremath{\mathit{#1}}}  %

\begin{algorithm}[t]
\small
\caption{Advantage filtering}
\label{alg:advfilter}
\begin{algorithmic}[1]
\Require Initial model parameters $\Theta$, RL environment $\mathcal{E}$, EWMA decay $\beta = 0.25$.
\For{$k = 0$ to $K-1$}
\Comment{Training iteration $k$}
\State $B_{\text{exp}} \gets \Call{CollectRollouts}{\Theta, \mathcal{E}}$
\Comment{Experience buffer}
\vspace{4pt}
\State $\hat{A}^{(\cdot)}_{\text{GAE}} \gets \Call{\textbf{GAE}}{B_{\text{exp}}, \Theta, \gamma, \lambda}$
\Comment{GAE advantages}
\vspace{4pt}
\State $A_{\text{max}} \gets \underset{t \in B_{\text{exp}}}{\max} \left| \hat{A}^{(t)}_{\text{GAE}} \right|$
\Comment{Max abs. advantage}
\vspace{4pt}
\State $\bar{A}_{\text{max}} \gets \mathbbm{1}_{k=0} A_{\text{max}} + \mathbbm{1}_{k>0} \left( \beta A_\text{max} + (1 - \beta) \bar{A}_{\text{max}} \right)$

\vspace{4pt}
\State Filtering threshold: $\eta \gets 0.01 \, \bar{A}_{\text{max}}$
\vspace{4pt}
\State $B_{\text{filtered}} \gets \Call{Filter}{B_{\text{exp}}, \left| \hat{A}^{(\cdot)}_{\text{GAE}} \right| < \eta }$
\vspace{4pt}
\State $\Theta \gets \Call{\textbf{PPO}}{B_{\text{filtered}}, \Theta}$
\EndFor
\end{algorithmic}
 \end{algorithm}

Empirically, we filter out on average $\sim80\%$ of all samples throughout training (over 90\% in the early epochs). This significantly increases learning throughput as we avoid computing gradients that contribute minimally to the overall parameter update. With the proposed adaptive threshold $\eta$ we observe a $2.3$-fold increase in training throughput, from \num{0.53} to \num{1.2} million steps per second. Our experiments suggest that application of this technique not only accelerates learning, but also yields more robust policies (\eg see Carla LAV results in \cref{fig:ablations_lowadv}). We hypothesize that advantage filtering could be beneficial across various RL setups, especially where data collection is cheaper than gradient calculation (\eg training LLMs on synthetic data).

Previously~\citet{tao2021repaint} explored an experience filtering approach based on the fixed advantage threshold, selecting a subset of maximally informative samples from the teacher policy in the context of transfer learning. We further develop this idea and propose a general adaptive method compatible with a broad class of reinforcement learning applications.

\section{Neural network architecture}
\label{sec:policy_arch}

Our actor and critic are parameterized by virtually identical compact neural networks with $3$ million parameters each ($6$ million trainable parameters combined). At the high level, the architecture consists of a fully-connected backbone ($[1024 \times 1024 \times 1024]$ MLP) which receives concatenated feature embeddings and outputs either a distribution over actions or a scalar value estimate (see Fig. \ref{fig:technical}d).

Observations represented by simple feature vectors ($S^{(t)}$, $G^{(t)}$, $C_{\text{reward}}$, \textit{etc.}) are trivially encoded by smaller fully-connected networks, \eg: $f^{(t)}_S = \texttt{MLP}\left( S^{(t)} \right)$.

Observations containing multiple features per agent $\left( W^{(t)}_{\text{lane}}, W^{(t)}_{\text{boundary}}, W^{(t)}_{\text{stop}}, A^{(t)} \right)$ are represented by sets of feature vectors and require permutation-invariant encoders~\cite{Zaheer2017}. For each feature type we employ a small fully-connected network to encode each element in a set (\eg an individual map feature in $W^{(t)}_{\text{lane}}$) followed by channel-wise maxpooling layer. Outputs of all maxpooling encoders are then concatenated.

Observation sets $W^{(t)}$ and $A^{(t)}$ represent some of the largest data structures in our computation graph containing over $10^8$ individual feature vectors per inference step across all agents. Storing all of these observations in the PPO's rollout buffer would be prohibitively expensive. Instead, we store only the world states $\mathbf{s}^{(t)}$ and reconstruct most of the observations as needed for each training minibatch.

In addition to that, for the largest feature sets $W_{\text{lane}}$ and $W_{\text{boundary}}$ we omit a fraction of input features (akin to the dropout regularization technique). We randomly drop 40\% of $W_{\text{boundary}}$ and 50\% of $W_{\text{lane}}$ features during training for each agent. This \textit{feature dropout} approach allows us to fit the full \ours training system on \SI{40}{\giga\byte} A100 GPUs, while additionally modelling potential sensor noise, increasing robustness of the policy and preventing overfitting.

\section{Benchmark evaluation} \label{sec:supplement_eval}

For each benchmark, we use the benchmark-provided simulation infrastructure, translate observations into \ours, and run \ours as a co-simulator.
For Waymax and CARLA, we co-simulate a single \ours step for each evaluation step.
nuPlan additionally requires trajectory predictions at each step.
At each time-step, we roll out an entire scenario using \ours policies for all agents for $0.3$ seconds, and use the subsequent observed trajectory from the driver as its trajectory.

\subsection{nuPlan benchmark evaluation}
\label{sec:nuplan_details}

The nuPlan benchmark consists of a training, validation and held out test set built from thousands of hours of data collected from Las Vegas, Boston, Pittsburgh, and Singapore. However, in our work we \emph{never use data from the training set}. The benchmark provides three different challenges, an open-loop (OL) Challenge 1, closed-loop non-reactive (CL-NR) Challenge 2, and closed-loop reactive (CL-R) Challenge 3. We focus only on challenge 3 because it represents the most realistic setting with both the driver and the traffic vehicles being closed loop. The reactive traffic vehicle agents are controlled by the Intelligent Driver Model (IDM)~\cite{kesting2010enhanced} with their initial placement taken from real traffic distributions. However pedestrians are non-reactive and follow their logged trajectory. After initialization, the IDM agent will select a series of lanes (independent of the log vehicles trajectory) and follow a centerline lane path provide by nuPlan’s map. Longitudinal acceleration control comes from the IDM equation.

\begin{equation*}
    \frac{dv}{dt} = a \left( 1 - \left( \frac{v}{v_0} \right)^\delta - \left( \frac{s^*}{s} \right)^2 \right)
\end{equation*}

With acceleration limit $a$, desired speed $v_0$, current speed $v$, distance to lead agent $s$, safety margin $s^*$, and exponent $\delta$ is usually set to $4$. The agent defaults to acceleration $a$ unless it’s close to desired $v_0$ or close to an object in its path. In this challenge, there is less diversity in behavior, but it has more realistic vehicle interactions when the driver deviates from the logged vehicle trajectory.

Due to the nuPlan online leaderboard evaluation servers no longer being accessible, we evaluate \ours on the Val14 benchmark, which has been shown to be a good proxy for the leaderboard evaluation and has other publicly reported results~\cite{dauner2023parting}.

The \ours policy achieves state of the art results zero-shot on the Challenge 3 closed loop score. The closed-loop score is measured from 0 to 100 and is a weighted evaluation composed of at-fault collisions, drive-able area compliance, driving direction compliance, progress towards goal, time-to-collision bounds, speed limit compliance, and comfort. Full results are shown in \cref{tab:nuplan_results}. We are able to compute the component scores for the PDM-Hybrid entrant and simply present the total score for the other entrants.
Finally, we categorized all videos of nuPlan’s collisions and we provide them in the supplementary material.

\subsection{CARLA benchmark evaluation}
\label{sec:carla_details}

\citet{jaeger2023hidden} implemented a CARLA expert driver that serves as a teacher for a trained perception-based model.
To our best knowledge, the presented expert manifests the best scores on CARLA benchmarks ever reported in the literature. This expert makes use of privileged information such as exact positions of all traffic participants and perfect map information. Since the default CARLA leaderboard tracks do not provide such information, Jaeger et al.\ adapt the leaderboard code accordingly. We adopt this procedure and extract this information in a similar way. Moreover, we identified a bug in CARLA that can lead to faulty pedestrian states when a pedestrian collides with other traffic participants. This bug can result in erroneous collision infractions; see video in the supplementary material. We obviate this bug by removing a pedestrian actor from the scene in case of an event. To ensure the same test bed for all competing approaches, we run all of them in our setup and report the resulting scores along with the scores reported by the authors.
In addition to the expert in~\citet{jaeger2023hidden}, we compare to the CARLA autopilot and another expert from~\citet{chitta2023transfuser}. Since simulations in CARLA contain a large amount of randomness, \cref{tab:carla_results} presents the mean and standard deviations of the evaluation results for 3 different runs.

\subsection{Waymax benchmark evaluation}
\label{sec:waymax_details}

To run the Waymax simulator, we wrote a distributed scenario runner to evaluate on the full Waymo Open Motion Dataset (WOMD) $1.2.0$ validation set consisting of $\num{44097}$ scenarios each \SI{8}{\second} long running at \SI{10}{\hertz}. To match the results from Table 3 in~\citet{gulino2023waymax}, for our \texttt{DatasetConfig} we use \texttt{max\_num\_objects=128}, and use the default \texttt{IDMRoutePolicy} settings. We also applied this  patch, \url{https://github.com/waymo-research/waymax/pull/54}, to fix IDM agent issues on top of commit \texttt{720f9214a} in the public Waymax Github repo. For goals used for the \ours policy we use the final location of the expert logged trajectory as an intermediate goal, as well as two more goals projected out at \SI{50}{\meter} and \SI{100}{\meter} and snapped to the nearest lane center.

We use the provided agent overlap, off road rate, log divergence, and SDC kinematic infeasibility metrics provided by Waymax to calculate our results. However, due to the Waymax route information not being released at the time of this writing, we report a modified metric for \texttt{Route Progress Ratio}. We calculate it by dividing the distance the \ours policy drove by the expert logged trajectory \textit{per scenario}, and clamp to 100\% if the expert drove less than \SI{1}{\meter}. We also report an additional progress metric, \texttt{Total Distance Driven Ratio}, which is simply the total distance the \ours policy drove over all scenarios divided by the total the expert log drove.

Unlike Carla and nuPlan, Waymax does not have an aggregate score which allows for easier comparison and visualization of results. Because of this, we propose a simple aggregate score composed of a progress score (i.e. a clamped \texttt{Total Distance Driven Ratio}), collision rates, and off-road rates show in \cref{eq:waymax-score} below. This metric is what is reported in our figures when referring to a single Waymax score.

\begin{samepage}
\begin{align}
    \text{progress} &= \min\left(\frac{\sum\limits_{s\in\mathcal{S}} d_{s}}{\sum\limits_{s\in\mathcal{S}} d_{s}^{\text{ expert}}}, \quad 1\right)  \notag \\[10pt]
    \text{success} &= 1 - \frac{\sum\limits_{s\in\mathcal{S}} \mathbbm{1}_s^{\text{collision}} \times \mathbbm{1}_s^{\text{off-road}} }{|\mathcal{S}|} \notag \\[8pt]
    \text{score} &= \text{progress} \times \text{success} \label{eq:waymax-score}
\end{align}
\end{samepage}

Where $\mathcal{S}$ is the set of scenarios, $d_s$ is the distance the policy under test drove in scenario $s$, $d_s^{\text{ expert}}$ is the distance the expert policy drove in scenario $s$, $\mathbbm{1}_s^{\text{collision}}$ is the indicator function for if scenario $s$ contained a collision, and $\mathbbm{1}_s^{\text{off-road}}$ is the indicator function for if scenario $s$ contained an off-road event.

\cref{tab:waymax_results} presents Waymax results.

\subsection{Waymo Open Sim-Agents Challenge} %
\label{sec:simagents}

Waymo introduced their Open Sim Agents Challenge (WOSAC) in 2023 to promote the development of intelligent, interactive simulation agents capable of exhibiting a diversity of behaviors in response to decisions made by the autonomous vehicle~\cite{montali2023waymo}.
In WOSAC, driving scenarios are characterized by 9.1 second sequences of recorded tracks of road users derived from the Waymo Open Motion Dataset (WOMD)~\cite{ettinger2021large}, stored at 10 Hz, where the first $\SI{1.1}{\second}$ of history is used to form an initial context.
Models are evaluated by their ability to accurately reproduce the remaining $\SI{8}{\second}$ of the scenario for up to 128 agents.

Despite being designed for imitative traffic modeling, we use WOSAC to evaluate \ours in real-world multi-agent settings and demonstrate the policy's ability to control a diversity of road users while avoiding undesired behavior.
Following the evaluation of other benchmarks, we evaluate a fully trained \ours policy zero-shot on WOSAC without fine-tuning.
We use the provided $\SI{1.1}{\second}$ history to conditionally sample from the map distribution of possible goal locations, filtering out positions that are either unreachable or are trivially reachable.
Proposed goal locations are retained proportional to the \ours agent's value estimate from the provided initial position of each actor.

Behavior of all vehicles and bicycles is modeled using a \ours policy, while pedestrians are controlled with an IDM-like policy based on the provided initial velocity and heading.
If a collision with a vehicle is forecast within a $\SI{0.5}{\second}$ horizon, the pedestrian actor stops moving forward until a collision is no longer imminent.
To account for sensor noise in recorded data, we add random Gaussian noise to all generated actor positions.
We achieved a 2-point improvement in the overall metametric using $\sigma=0.0125$ when compared to rollouts without the added noise.

Results of our zero-shot evaluation of \ours to WOSAC are reported in \cref{tab:wosac_table}, using the Waymo-defined validation set to allow for detailed analysis in \cref{tab:wosac_collisions_and_offroad}.
In addition to \ours we include a comparison to a policy where all actors are fixed to be stationary, establishing a lower bound on generated behaviors by the \ours policy.
We also include the reported performance for the top methods included on the public 2023 WOSAC leaderboard as well as a set of baseline approaches reported in~\citet{montali2023waymo}, nearly all based around supervised autoregressive modeling.

We see that the \ours policy approaches the performance of expert models, despite having never been exposed to any map or human data recorded in WOMD during training.
Notably, \ours offers strong performance for collision and off-road metrics, out-performing most approaches.
\ours also maintains respectable performance among acceleration metrics, a consequence of producing kinematically feasbile and smooth rollouts following the dynamics model the policy was trained with.
With \ours, we are able to control multiple actors within real-world scenarios without colliding or driving off-road, producing naturalistic driving.
The scoring metrics used for WOSAC are likelihood measures of the ground truth behavior within the distribution induced by generated actor trajectories.
To achieve idealized performance for the likelihood of collision and off-road behavior, it is expected that the generated trajectories collide or drive off-road whenever ground truth actors are assessed to do so.
Because \ours explicitly penalizes these outcomes during training we cannot expect our policy to substantively improve over the reported scores in \cref{tab:wosac_table} for these metrics.
We characterize the collision and off-road behavior of \ours within WOSAC in \cref{tab:wosac_collisions_and_offroad}, conditioned on actor type and whether the ground truth trajectories collide or go off-road.

\section{Analysis}

\subsection{Reward conditioning analysis}

To capture a continuum of driving styles in \ours we vary reward parameters for our agents and then allow the policy to condition on them. In total we condition on 12 different parameters used in our reward shown in \cref{tab:reward_rand}.

To better understand how these reward parameter conditions affect \ours policy behavior, we attempt to find which parameters can explain the most variation in trajectories. We do this by sampling 200 random reward parameters within the training distribution range, roll out each of them on various scenarios in nuPlan, then cluster trajectories to find which conditioned reward parameter provides the most mutual information on the clusters. Because single reward parameter will typically dominate the variation seen in the trajectories, we repeat the analysis again but with the previous dominant parameter fixed. We repeat this procedure for the top 6 parameters with the results shown in \cref{fig:conditional-analysis}.

The order of parameters with highest mutual information with the clusters is $\alpha_{\text{center-bias}}$ followed by $\delta_{\text{goal}}$, $\alpha_{\text{l-center}}$, $\alpha_{\text{comfort}}$, $\alpha_{\text{l-align}}$, and $\alpha_{\text{vel-align}}$. As seen in \cref{fig:conditional-analysis}, the variation is interpretable and relatable to a human driver. For $\alpha_{\text{center-bias}}$, the policy will pick the left lane if it's biased towards the right (so that it can be further from the boundary) and vice versa for the right lane. For $\delta_{\text{goal}}$, the policy will change lanes to get closer to the goal when the max distance is small enough. For $\alpha_{\text{l-center}}$, if this value is high enough, the policy prefers to stay centered in its lane to not incur the extra cost of a lane change, despite the additional progress it could make. For $\alpha_{\text{comfort}}$, the policy cares less about high accelerations and jerks when this penalty is closer to zero, so will take more aggressive turns to stay within its lane and receive centering rewards. For $\alpha_{\text{l-align}}$, the policy incurs less reward for staying oriented with its lane if this value is low, and therefore if this value is low enough the driver is willing to U-turn to get to its goal behind it. Finally for $\alpha_{\text{vel-align}}$, we are rewarded for staying centered in our lane at higher speeds, and therefore when this value is low the policy takes a wider turn to maximize comfort.

By allowing our policy to condition on a range of reward parameters, we solve a few problems at once while keeping the simplicity of a single policy. First, it introduces a wide variety of agent behaviors that makes our policy more robust. One reason is because the policy now must handle both the hidden goals and hidden behaviors of other agents. Due to this, the \ours policy learns to handle vehicles that favor different positioning within the lane, drive aggressively, lane change more often, plus combinations of the above. The second benefit is that the optimal driving style could be captured in our conditioning range. So as an additional alignment step, the reward conditions can be tuned to reduce the error between the \ours policy and an ideal logged driver. This method would then allow us to create both a chaotic self-play training environment to increase robustness, and a safe and comfortable driving controller, all with the same policy.

\subsection{Ablation studies}

We perform a series of experiments to understand how different algorithmic choices influence the final performance of the \ours agent using a reduced compute budget of 660 GPU-hours ($\sim30\%$ of the final training run's budget).

\cref{fig:ablations_lowadv} demonstrates the effect of the \textit{advantage filtering} technique described earlier. %
Using this method we filter out samples that have near zero contribution to PPO's policy and critic losses, thus focusing on rare events at the tails of data distribution. While we expect this technique to improve the wall time performance of the algorithm, to our surprise we also see qualitative difference at convergence. The version of \ours without filtering plateaus substantially lower on the Carla LAV benchmark and is unable to reach up to 5\% of goals in time leading to timed out episodes in the self-play evaluation mode.

\cref{fig:ablations_combined} demonstrates the influence of various algorithmic features on zero-shot benchmark performance. Here, we quantify the error rate across multiple benchmarks; specifically, we measure the additional percentage points needed to achieve a perfect score on each benchmark.

\subsection{Long-form evaluation in self-play}

To evaluate robustness of our agent in self-play, we devise a version of \ours training environment tuned for long-form evaluation. We remove dynamics noise, set the number of drivers to $N_a = 50$, and slightly modify the initialization procedure to ensure minimal clearance between all vehicles at $t=0$. We remove all traffic light randomization and simply use CARLA's default traffic light system to promote conventional interactions at intersections. We additionally modify drivers' observations to increase their perceived vehicle size by \SI{10}{\centi\meter} on each size.

We observe that at decision-making frequency of \SI{15}{\hertz}, \ours agent conditioned for safe and conservative driving (see \cref{tab:megadrive_big_eval}) exhibits nearly accident-free driving, covering on average $3$ million km before encountering any collision or going off-road (\cref{fig:self-play-robustness}).

\clearpage
\setcounter{figure}{0}
\renewcommand\thefigure{A\arabic{figure}}
\setcounter{table}{0}
\renewcommand\thetable{A\arabic{table}}

\begin{table*}[h]
    \centering
    \small
    \begin{tabular}{@{}lcc@{}}
        \toprule
        \textbf{Parameter} & \textbf{Value} \\
        \midrule
        Total num. of maps w/ affine transformations & 128 \\
        Number of agents per sim. & $U\{1, 150\}$\\
        Number of sims per GPU & $\num{4800}$ \\
        Max number of agents per GPU & $\num{720000}$ \\
        GPUs used & $8 \times$ Nvidia A100 (\SI{40}{\giga\byte}) \\
        Total GPU hours used in training & 1900  \\
        Total number of sims & $\num{38400}$ \\
        Total number of agents & $\num{5760000}$ \\
        Throughput: steps per second & $\sim \num{1200000}$ \\
        Throughput: km driven per second & $\sim \num{1850}$ \\
        Training cost per million km driven & $\sim\, $\SI{5}[\$]{USD} \\
        Total simulated distance driven & $\SI{1600000000}{\kilo\meter}$ \\
        Total number of simulated state transitions & $\num{1000000000000}$\\

        \bottomrule
    \end{tabular}%}
    \caption{Setup and scale of \ours training.}
    \label{tab:scale}
\end{table*}

\begin{table*}
\centering
\small
\resizebox{\linewidth}{!}{%
\begin{tabular}{@{}r@{}l@{\hspace{5mm}}r@{}l@{}}
\toprule
\multicolumn{2}{c}{\textbf{Reward}} & \multicolumn{2}{c}{\textbf{Training distribution}} \\   \midrule \addlinespace[5pt]

\multirow{2}{*}{$R_{\text{goal}} $} & \multirow{2}{*}{$=\mathbbm{1}_{(\lvert\lvert x - g \rvert \rvert < \delta_{\text{goal}} \,\land\, (\mathbbm{1}_{\text{waypoint}} \lor \lvert v \rvert < v_\text{goal}))}$}
& $\delta_{\text{goal}}$ & $\sim U(2, 12)$ \\
& & $v_{\text{goal}}$ & $= 3$
\\ [5pt]  \midrule \addlinespace[5pt]

$R_{\text{collision}} $&$=  - \left( \alpha_\text{collision} + 0.1 \lvert v \rvert \right)\mathbbm{1}_\mathrm{collision}$ &
$\alpha_{\text{collision}} $ & $\sim U(0, 3)$
\\ [5pt]  \midrule \addlinespace[5pt]

$R_{\text{off-road}} $&$= -\alpha_\text{boundary} \mathbbm{1}_\text{boundary}$& $\alpha_{\text{boundary}}$ & $\sim U(0, 3)$
\\ [5pt]  \midrule \addlinespace[5pt]

$R_{\text{comfort}} $&$= -\alpha_\text{comfort} \left( \mathbbm{1}_{\lvert a_\text{long} \rvert > 3} + \mathbbm{1}_{\lvert a_\text{lat} \rvert > 3} +
    \mathbbm{1}_{\lvert \dot{a}_\text{long} \rvert > 5 \, \lor \, \lvert \dot{a}_\text{lat} \rvert > 5} \right)$ &
$\alpha_{\text{comfort}}$&$\sim U(0.0, 0.1)$
\\ [5pt]  \midrule \addlinespace[5pt]

\multirow{2}{*}{$R_{\text{l-align}}$} & \multirow{2}{*}{$=\alpha_{\text{l-align}} \Delta t \left(\min\left(\cos\left(\theta_f\right), 0\right) + \alpha_{\text{vel-align}}\min\left(\cos\left(\theta_f\right) * v, 0 \right)+ 0.0025 \left(1 - \frac{\lvert \theta_f \rvert}{\pi/2}\right)\right)$}&
$\alpha_{\text{l-align}}$ & $\sim U(\mynum{0.00025}, \mynum{0.025})$ \\
& & $\alpha_{\text{vel-align}}$ & $\sim U(0, 1)$ \\ [5pt]
\midrule \addlinespace[5pt]

\multirow{2}{*}{$R_{\text{l-center}}$} &\multirow{2}{*}{$= -\alpha_\text{l-center} \Delta t\left(\mathbbm{1}_{\cos\left(\theta_f\right) > 0.5}*\lvert x_f - \alpha_{\text{center-bias}} \rvert - \frac{0.05}{\exp \left(\lvert x_f - \alpha_{\text{center-bias}} \rvert - 0.5\right)}\right)$}& $\alpha_{\text{l-center}} $&$\sim U(\mynum{0.00025}, \mynum{0.0075})$ \\
& & $\alpha_{\text{center-bias}}$ & $ \sim U(-0.5, 0.5)$
\\ [5pt]  \midrule \addlinespace[5pt]

$R_{\text{velocity}} $&$= \alpha_{\text{velocity}} \, \Delta t \, \max\left( \cos \left(\theta_f \right), 0.0\right) \, \mathbbm{1}_{\lvert v \rvert > 2.5}$ & $\alpha_{\text{velocity}} $&$= \mynum{0.0025}$
\\ [5pt]  \midrule \addlinespace[5pt]

$R_{\text{reverse}} $&$= - \alpha_\text{reverse} \; \Delta t \, \mathbbm{1}_{v < 0}$ &
$\alpha_{\text{reverse}}$&$\sim U(\mynum{0.00025}, \mynum{0.0075})$
\\ [5pt]  \midrule \addlinespace[5pt]

$R_{\text{stop-line}} $&$= - \alpha_\text{stop-line} \, \mathbbm{1}_\text{stop-line-violation}$
& $\alpha_{\text{stop-line}}$&$\sim U(0, 1)$
\\ [5pt]  \midrule \addlinespace[5pt]

$R_{\text{timestep}} $&$=  - \left(\alpha_{\text{timestep}} \Delta t  \right) \, \mathbbm{1}_{\lvert v \rvert > 0  \,\lor \, \lvert a \rvert > 0}$                               & $\alpha_{\text{timestep}} $ & $= \mynum{0.000025}$       \\ [5pt]

\bottomrule
\end{tabular}}
\caption{Training distribution for each of the reward components.}
\label{tab:reward_rand}
\end{table*}

\begin{table*}[ht]
\centering
\small
\begin{minipage}[b]{0.47\linewidth} %
\centering
    \begin{tabular}{lc}
    \toprule
    \textbf{Parameter} & \textbf{Value} \\
    \midrule
     Training batch size & $\num{256000}$ \\
     Batch size per GPU & $\num{32000}$ \\
     Rollout length & $\num{128}$ \\
     Num. PPO epochs & 3 \\
     Discount factor $\gamma$ & 0.999 \\
     $\lambda_\text{GAE}$ & 0.95 \\
     Max. episode length & 1200 steps (\SI{360}{\second}) \\
     PPO clipping ratio & 0.2 \\
     Value function clipping & None \\
     Initial LR $\alpha^{(0)}$ & $\num{5e-4}$\\
     LR schedule & Cosine\\
     Entropy coefficient & 0.01 \\
     Value loss coefficient & 0.5 \\
     Max grad. norm & 0.5 \\
     Advantage normalization & Enabled \\
     Adv. filtering threshold $\eta$ & $0.01 \, \bar{A}_{\text{max}}$ (Alg. \ref{alg:advfilter}) \\
     Inference \& training precision & 16-bit AMP \\
     Model weights initialization & Orthogonal, zero bias \\
     \bottomrule
    \end{tabular}%}
    \caption{RL algorithm settings and hyperparameters used during training.}
    \label{tab:ppo_params}
\end{minipage}
\hfill
\begin{minipage}[b]{0.47\linewidth} %
\centering
    \begin{tabular}{lc}
        \toprule
        \textbf{Setting/Parameter} & \textbf{Value} \\
        \midrule
        Vehicle length & $l\sim U(2, 5.5)$ \\
        Vehicle width & $w\sim U(1.5, 2.5)$ \\
        Maximum speed & \SI{20}{\meter\per\second} \\
        Timestep $\Delta t$ & \SI{0.066}{\second} \\
        Episode length & 9000 steps (\SI{600}{\second}) \\
        Agents per sim $N_a$ & $\num{50}$ \\
        Observed agents $N_o$ & up to $\num{40}$ closest \\
        \midrule
        $\delta_{\text{goal}}$ & $\SI{10}{\meter}$ \\
        $v_{\text{goal}}$ & $\SI{3}{\meter\per\second}$  \\
        $\alpha_{\text{collision}}$ & $\num{3.0}$ \\
        $\alpha_{\text{boundary}}$ & $\num{3.0}$ \\
        $\alpha_{\text{comfort}}$ & $\num{0.05}$ \\
        $\alpha_{\text{l-align}}$ & $\mynum{0.025}$ \\
        $\alpha_{\text{vel-align}}$ & $1.0$ \\
        $\alpha_{\text{l-center}}$ & $\mynum{0.00375}$\\
        $\alpha_{\text{center-bias}}$ & 0.0\\
        $\alpha_{\text{velocity}}$ & $\mynum{0.0025}$\\
        $\alpha_{\text{reverse}}$ & $\mynum{0.005}$\\
        $\alpha_{\text{stop-line}}$ & 1.0\\
        $\alpha_{\text{timestep}}$ & $\mynum{0.000025}$ \\
        \bottomrule
    \end{tabular}
    \caption{\ours settings for long-form evaluation in self-play.}
    \label{tab:megadrive_big_eval}
\end{minipage}
\end{table*}

\begin{table*}[h]
\centering
\resizebox{\linewidth}{!}{
\begin{tabular}{l|ccccccccc}
\toprule
Method & Score $\uparrow$ & Ego Progress$\uparrow$ & No AF-Collisions $\uparrow$ & Comfort $\uparrow$ & TTC in bounds $\uparrow$ & Driving Direction
 $\uparrow$ & Speed Limit $\uparrow$ & Drivable Area $\uparrow$ & Ego Making Progress $\uparrow$ \\
\midrule
\multicolumn{9}{l}{\textit{Val14 Benchmark}} \\
Urban Driver~\cite{scheel2021urban} & 50 & - & - & - & - & - & - & - & - \\
GC-GPP~\cite{hallgarten2023prediction} & 55 & - & - & - & - & - & - & - & - \\
PlanCNN~\cite{renz2022plant} & 72 & - & - & - & - & - & - & - & - \\
IDM~\cite{Treiber_2000} & 77 & - & - & - & - & - & - & - & - \\
PDM-Hybrid~\cite{dauner2023parting} & 92.1 & 90.2 & 98.1 & 94.8 & 93.5 & 99.9 & 99.8 & 99.5 & 99.1 \\
Diffion-ES~\cite{yang2024diffusion} & 92.2 & 91.2 & 97.7 & 93.4 & 93.8 & 100.0 & 99.7 & 99.6 & 99.2 \\
\ours (ours) & 93.8\textcolor{gray}{$\pm$0.11} & 93.6\textcolor{gray}{$\pm$0.06} & 98.4\textcolor{gray}{$\pm$0.09} & 96.4\textcolor{gray}{$\pm$0.27} & 93.8\textcolor{gray}{$\pm$0.23}& 99.6\textcolor{gray}{$\pm$0.05} & 99.9\textcolor{gray}{$\pm$0.00} & 99.7\textcolor{gray}{$\pm$0.04} & 99.0\textcolor{gray}{$\pm$0.07} \\
\bottomrule
\end{tabular}}
\caption{Results on Val14~\cite{dauner2023parting} nuPlan benchmark.}
\label{tab:nuplan_results}
\end{table*}

\begin{table*}[h]
\centering
\resizebox{\linewidth}{!}{%
\begin{tabular}{l|ccc|ccccccccc}
\toprule
Method & DS$\uparrow$ & RC$\uparrow$ & IP$\uparrow$ & Ped$\downarrow$ & Veh$\downarrow$ & Lay$\downarrow$ & Red$\downarrow$ & Stop$\downarrow$ & Off$\downarrow$ & Dev$\downarrow$ & TO$\downarrow$ & Block$\downarrow$ \\
\midrule
\multicolumn{13}{l}{\textit{CL1 Testing Routes (\url{leaderboard.carla.org/get_started_v1})}} \\
CARLA Agent (our run) & 29 \textcolor{gray}{$\pm$0} & 41 \textcolor{gray}{$\pm$0} & 0.73 \textcolor{gray}{$\pm$0.03} & 0.03 \textcolor{gray}{$\pm$0.00} & 0.36 \textcolor{gray}{$\pm$0.11} & 0.00 \textcolor{gray}{$\pm$0.00} & 0.28 \textcolor{gray}{$\pm$0.02} & 0.03 \textcolor{gray}{$\pm$0.02} & 0.00 \textcolor{gray}{$\pm$0.00} & 0.44 \textcolor{gray}{$\pm$0.02} & 0.01 \textcolor{gray}{$\pm$0.01} & 0.15 \textcolor{gray}{$\pm$0.00} \\
Expert from~\citet{jaeger2023hidden} (our run) & 90 \textcolor{gray}{$\pm$0} & 96 \textcolor{gray}{$\pm$1} & 0.94 \textcolor{gray}{$\pm$0.01} & 0.00 \textcolor{gray}{$\pm$0.00} & 0.06 \textcolor{gray}{$\pm$0.01} & 0.00 \textcolor{gray}{$\pm$0.00} & 0.06 \textcolor{gray}{$\pm$0.03} & 0.00 \textcolor{gray}{$\pm$0.00} & 0.00 \textcolor{gray}{$\pm$0.00} & 0.00 \textcolor{gray}{$\pm$0.00} & 0.05 \textcolor{gray}{$\pm$0.02} & 0.05 \textcolor{gray}{$\pm$0.00} \\
\ours (ours) & 93 \textcolor{gray}{$\pm$1} & 97 \textcolor{gray}{$\pm$2} & 0.95 \textcolor{gray}{$\pm$0.01} & 0.00 \textcolor{gray}{$\pm$0.00} & 0.07 \textcolor{gray}{$\pm$0.01} & 0.00 \textcolor{gray}{$\pm$0.00} & 0.01 \textcolor{gray}{$\pm$0.01} & 0.00 \textcolor{gray}{$\pm$0.00} & 0.64 \textcolor{gray}{$\pm$0.44} & 0.01 \textcolor{gray}{$\pm$0.01} & 0.07 \textcolor{gray}{$\pm$0.02} & 0.00 \textcolor{gray}{$\pm$0.00} \\
\midrule
\multicolumn{13}{l}{\textit{LAV Benchmark~\cite{chen2022lav} (with adapted scenarios from~\citet{jaeger2023hidden})}} \\
CARLA Agent (our run) & 9 \textcolor{gray}{$\pm$2} & 58 \textcolor{gray}{$\pm$0} & 0.18 \textcolor{gray}{$\pm$0.04} & 0.22 \textcolor{gray}{$\pm$0.00} & 0.74 \textcolor{gray}{$\pm$0.43} & 0.00 \textcolor{gray}{$\pm$0.00} & 3.99 \textcolor{gray}{$\pm$0.10} & 0.58 \textcolor{gray}{$\pm$0.21} & 0.00 \textcolor{gray}{$\pm$0.00} & 0.87 \textcolor{gray}{$\pm$0.00} & 0.00 \textcolor{gray}{$\pm$0.00} & 0.00 \textcolor{gray}{$\pm$0.00} \\
Expert from~\citet{jaeger2023hidden} & 94 & 95 & 0.99 & 0.00 & 0.02 & 0.00 & 0.02 & 0.00 & - & 0.00 & 0.00 & 0.08 \\
Expert from~\citet{jaeger2023hidden} (our run) & 92 \textcolor{gray}{$\pm$9} & 95 \textcolor{gray}{$\pm$7} & 0.98 \textcolor{gray}{$\pm$0.02} & 0.00 \textcolor{gray}{$\pm$0.00} & 0.04 \textcolor{gray}{$\pm$0.05} & 0.00 \textcolor{gray}{$\pm$0.00} & 0.02 \textcolor{gray}{$\pm$0.03} & 0.00 \textcolor{gray}{$\pm$0.00} & 0.00 \textcolor{gray}{$\pm$0.00} & 0.00 \textcolor{gray}{$\pm$0.00} & 0.09 \textcolor{gray}{$\pm$0.07} & 0.07 \textcolor{gray}{$\pm$0.07} \\
\ours (ours) & 99 \textcolor{gray}{$\pm$1} & 99 \textcolor{gray}{$\pm$1} & 1.00 \textcolor{gray}{$\pm$0.00} & 0.00 \textcolor{gray}{$\pm$0.00} & 0.00 \textcolor{gray}{$\pm$0.00} & 0.00 \textcolor{gray}{$\pm$0.00} & 0.00 \textcolor{gray}{$\pm$0.00} & 0.00 \textcolor{gray}{$\pm$0.00} & 0.00 \textcolor{gray}{$\pm$0.00} & 0.00 \textcolor{gray}{$\pm$0.00} & 0.02 \textcolor{gray}{$\pm$0.03} & 0.00 \textcolor{gray}{$\pm$0.00} \\
\midrule
\multicolumn{13}{l}{\textit{Longest6 Benchmark~\cite{chitta2023transfuser}}} \\
CARLA Agent (our run) & 7 \textcolor{gray}{$\pm$1} & 54 \textcolor{gray}{$\pm$0} & 0.25 \textcolor{gray}{$\pm$0.01} & 0.31 \textcolor{gray}{$\pm$0.00} & 1.50 \textcolor{gray}{$\pm$0.21} & 0.00 \textcolor{gray}{$\pm$0.00} & 1.84 \textcolor{gray}{$\pm$0.06} & 0.10 \textcolor{gray}{$\pm$0.02} & 0.00 \textcolor{gray}{$\pm$0.00} & 0.43 \textcolor{gray}{$\pm$0.00} & 0.02 \textcolor{gray}{$\pm$0.01} & 0.04 \textcolor{gray}{$\pm$0.00} \\
Expert from~\citet{chitta2023transfuser} & 77 \textcolor{gray}{$\pm$2} & 89 \textcolor{gray}{$\pm$1} & 0.86 \textcolor{gray}{$\pm$0.03} & 0.02 & 0.28 & 0.01 & 0.03 & 0.00 & 0.00 & 0.00 & 0.08 & 0.13 \\
Expert from~\citet{jaeger2023hidden} & 81 \textcolor{gray}{$\pm$3} & 90 \textcolor{gray}{$\pm$1} & 0.91 \textcolor{gray}{$\pm$0.04} & 0.01 & 0.21 & 0.00 & 0.01 & - & - & 0.00 & 0.07 & 0.09 \\
Expert from~\citet{jaeger2023hidden} (our run) & 83 \textcolor{gray}{$\pm$1} & 94 \textcolor{gray}{$\pm$2} & 0.88 \textcolor{gray}{$\pm$0.01} & 0.00 \textcolor{gray}{$\pm$0.00} & 0.20 \textcolor{gray}{$\pm$0.03} & 0.00 \textcolor{gray}{$\pm$0.00} & 0.04 \textcolor{gray}{$\pm$0.01} & 0.02 \textcolor{gray}{$\pm$0.01} & 0.00 \textcolor{gray}{$\pm$0.00} & 0.00 \textcolor{gray}{$\pm$0.00} & 0.09 \textcolor{gray}{$\pm$0.00} & 0.06 \textcolor{gray}{$\pm$0.02} \\
\ours (ours) & 92 \textcolor{gray}{$\pm$2} & 99 \textcolor{gray}{$\pm$1} & 0.93 \textcolor{gray}{$\pm$0.01} & 0.02 \textcolor{gray}{$\pm$0.00} & 0.08 \textcolor{gray}{$\pm$0.01} & 0.00 \textcolor{gray}{$\pm$0.00} & 0.04 \textcolor{gray}{$\pm$0.02} & 0.03 \textcolor{gray}{$\pm$0.01} & 0.24 \textcolor{gray}{$\pm$0.08} & 0.03 \textcolor{gray}{$\pm$0.02} & 0.05 \textcolor{gray}{$\pm$0.03} & 0.00 \textcolor{gray}{$\pm$0.00} \\
\bottomrule
\end{tabular}}
\caption{Results on CARLA benchmarks.
}
\label{tab:carla_results}
\end{table*}

\begin{table*}[h]
\centering
\resizebox{\linewidth}{!}{
\begin{tabular}{l|ccccccc}
\toprule
Method & Off-Road Rate (\%) $\downarrow$ & Collision Rate (\%) $\downarrow$ & Kinematic Infeasibility (\%)  $\downarrow$ & Log ADE (m) & Route Progress Ratio (\%) $\uparrow$ & Total Distance Driven Ratio (\%) $\uparrow$ & Score (\%) $\uparrow$\\
\midrule
Expert Demonstration~\cite{gulino2023waymax} & 0.32 & 0.61 & 4.33 & 0.00 & 100.00 & 100.00 & $\le$ 99.07  \\
\specialrule{0.01pt}{1pt}{1pt} %
Wayformer~\cite{gulino2023waymax} & 7.89 & 10.68 & 5.40 & 2.38 & 123.58 & - & $\le$
 81.43 \\
DQN~\cite{gulino2023waymax} & 3.74\textcolor{gray}{$\pm$0.90} & 6.50\textcolor{gray}{$\pm$0.31} & 0.00\textcolor{gray}{$\pm$0.0} & 9.83\textcolor{gray}{$\pm$0.48} & 177.91\textcolor{gray}{$\pm$5.67} & - & $\le$ 89.76\textcolor{gray}{$\pm$0.95}\\
BC~\cite{gulino2023waymax} & 1.11\textcolor{gray}{$\pm$0.2} & 4.59\textcolor{gray}{$\pm$0.06} & 0.00\textcolor{gray}{$\pm$0.0} & 2.26\textcolor{gray}{$\pm$0.2} & 129.84\textcolor{gray}{$\pm$0.98} & - & $\le$ 94.3\textcolor{gray}{$\pm$0.21} \\
\ours (ours) & 0.43\textcolor{gray}{$\pm$0.008} & 0.43\textcolor{gray}{$\pm$0.005} & 0.14\textcolor{gray}{$\pm$0.008} & 5.87\textcolor{gray}{$\pm$0.01} & 146.27\textcolor{gray}{$\pm$0.08} & 106.79\textcolor{gray}{$\pm$0.05} & 99.16\textcolor{gray}{$\pm$0.009} \\
\bottomrule
\end{tabular}}
\caption{Results on Waymax benchmarks when evaluated with IDM agents. For the \texttt{Score} of other agents, we assumed 100\% progress and mutually exclusive collision and offroad events.}
\label{tab:waymax_results}
\end{table*}

\begin{table*}[h]
\centering
\renewcommand{\arraystretch}{1.25}
\resizebox{\linewidth}{!}{
\begin{tabular}{l|ccccccccc||c}
\toprule
Method & Linear Speed & Linear Accel. & Ang. Speed  & Ang. Accel. & Dist. to Obj. & Collision  & TTC  & Dist. to Road Edge & Off-road  & {\bf Composite Metric} \\
\midrule
Expert Demo.*~\cite{montali2023waymo} & 0.5610 & 0.3300 & 0.5630 & 0.4890 & 0.4850 & 1.0000 & 0.8810 & 0.7130 & 1.0000 & 0.7220 \\
\specialrule{0.01pt}{1pt}{1pt} %
\rowcolor[HTML]{E6E6E6}
Random Agent* & 0.0020 & 0.0440 & 0.0740 & 0.1200 & 0.0000 & 0.0000 & 0.7340 & 0.1780 & 0.2870 & 0.1550 \\
\rowcolor[HTML]{E6E6E6}
Linear Extrapolation* & 0.1570 & 0.1190 & 0.0190 & 0.0350 & 0.2470 & 0.4110 & 0.7750 & 0.5020 & 0.4630 & 0.3240 \\
\rowcolor[HTML]{E6E6E6}
Stationary Policy & 0.0408 & 0.0612 & 0.4984 & 0.3195 & 0.0553 & 0.9466 & 0.7363 & 0.2408 & 0.8575 & 0.5007 \\
Joint-Multipath++*~\cite{wang2023joint} & 0.4340 & 0.2300 & 0.5150 & 0.4520 & 0.3450 & 0.5670 & 0.8120 & 0.6390 & 0.6820 & 0.5330 \\
PredSim* & 0.4051 & 0.2208 & 0.4958 & 0.4653 & 0.3441 & 0.7193 & 0.8027 & 0.6167 & 0.7519 & 0.5663 \\
Wayformer*~\cite{nayakanti2023wayformer} & 0.3310 & 0.0980 & 0.4130 & 0.4060 & 0.2970 & 0.8700 & 0.7820 & 0.5920 & 0.8660 & 0.5750 \\
SceneDMF*~\cite{guo2023scenedm} & 0.4315 & 0.2767 & 0.5230 & 0.4666 & 0.3678 & 0.7447 & 0.8128 & 0.6215 & 0.7392 & 0.5821 \\
MTR+++*~\cite{qian2023simple} & 0.4119 & 0.1066 & 0.4838 & 0.4365 & 0.3457 &  0.8630 & 0.7969 & 0.6545 & 0.8954 & 0.6077 \\
\rowcolor[HTML]{E6E6E6}
\ours (ours) & 0.2613 & 0.2534 & 0.5060 & 0.4756 & 0.3161 & 0.9470 & 0.8077 & 0.5445 & 0.9095 & 0.6190 \\
VPD-BP*~\cite{huang2024versatile} & 0.4751 & 0.2161 & 0.5358 & 0.4775 & 0.3908 & 0.8162 & 0.8234 & 0.6628 & 0.9010 & 0.6315 \\
MTR-E*~\cite{qian2023simple} & 0.4278 & 0.2353 & 0.5335 & 0.4753 & 0.3455 &  0.8774 & 0.7983 & 0.6541 & 0.9143 & 0.6348 \\
MVTE*~\cite{wang2023multiverse} & 0.4426 & 0.2218 &
0.5353 & 0.4810 & 0.3819 & 0.8943 & 0.8321 & 0.6641 & 0.9086 & 0.6448 \\
Trajeglish*~\cite{philion2023trajeglish} & 0.4504 & 0.1929 & 0.5382 & 0.4850 & 0.3869 & 0.9226 & 0.8369 & 0.6596 & 0.8864 & 0.6451 \\
InteractionFormer* & 0.4294 & 0.2394 & 0.5291 & 0.4780 & 0.3776 & 0.9591 & 0.8311 & 0.6464 & 0.9347 & 0.6587 \\
\bottomrule
\end{tabular}}
\caption{Per-component metrics, as defined and computed within WOSAC using the WOMD validation split. *-scores publicly reported on WOMD test set at \url{https://waymo.com/open/challenges/2023/sim-agents/} and from \citet{montali2023waymo}. All scores are derived using the 2023 metric definitions. Shaded rows correspond to approaches that do not use training data. We see that \ours, in zero-shot evaluation, is capable of producing effective driving performance that approaches expert policies specifically developed for imitative traffic modeling using the provided training data, even without having been shown WOMD data or maps previously.}
\label{tab:wosac_table}
\end{table*}

\begin{table*}[h]
\centering
\renewcommand{\arraystretch}{1.5}
\resizebox{\linewidth}{!}{
\begin{tabular}{lccc|ccc|ccc|ccc|ccc}
       \toprule
       \multicolumn{16}{c}{\textbf{Collisions}} \\
    \midrule
              & \multicolumn{3}{c}{{Overall}}         & \multicolumn{3}{c}{{Ego}}             & \multicolumn{3}{c}{{Vehicles}}        & \multicolumn{3}{c}{{Pedestrians}}     & \multicolumn{3}{c}{{Bicycles}}        \\
\multicolumn{1}{c}{Method} & \textit{Total} & \textit{$\neg$C} & \textit{C} & \textit{Total} & \textit{$\neg$C} & \textit{C} & \textit{Total} & \textit{$\neg$C} & \textit{C} & \textit{Total} & \textit{$\neg$C} & \textit{C} & \textit{Total} & \textit{$\neg$C} & \textit{C} \\
\midrule
Expert Demo.  & 0.0337  & 0.0000 & 1.0000 & 0.0048 & 0.0000 & 1.0000 & 0.0076 & 0.0000 & 1.0000 & 0.2986 & 0.0000 & 1.0000 & 0.0547 & 0.0000 & 1.0000  \\
\specialrule{0.01pt}{1pt}{1pt} %
\ours (ours) & 0.0322 & 0.0085 & 0.7082 & 0.0021 & 0.0006 & 0.3223 & 0.0080 & 0.0051 & 0.3700 & 0.2391 & 0.0029 & 0.7938 & 0.0291 & 0.0003 & 0.5268   \\
Stationary & 0.0220 & 0.0017 & 0.6015 & 0.0021 & 0.0007 & 0.2832 & 0.0035 & 0.0012 & 0.3513 & 0.2012 & 0.0003 & 0.6731 & 0.0242 & 0.0001 & 0.3556  \\
\bottomrule
\multicolumn{16}{l}{} \\
\multicolumn{16}{c}{\textbf{Off-road}} \\
\midrule
& \multicolumn{3}{c}{{Overall}}         & \multicolumn{3}{c}{{Ego}}             & \multicolumn{3}{c}{{Vehicles}}        & \multicolumn{3}{c}{{Pedestrians}}     & \multicolumn{3}{c}{{Bicycles}}        \\
\multicolumn{1}{c}{} & \textit{Total} & \textit{$\neg$O} & \textit{O} & \textit{Total} & \textit{$\neg$O} & \textit{O} & \textit{Total} & \textit{$\neg$O} & \textit{O} & \textit{Total} & \textit{$\neg$O} & \textit{O} & \textit{Total} & \textit{$\neg$O} & \textit{O} \\
\midrule
Expert Demo. & 0.1263 & 0.0000 & 1.0000 & 0.0127 & 0.0000 & 1.0000 & 0.0599 & 0.0000  & 1.0000 & 0.8513 & 0.0000 & 1.0000 & 0.3611 & 0.0000 & 1.0000 \\
\specialrule{0.01pt}{1pt}{1pt} %
\ours (ours) & 0.1105 & 0.0072 & 0.8252 & 0.0071 & 0.0003 & 0.5367 & 0.0409 & 0.0051 & 0.6033 & 0.8385 & 0.0014 & 0.9847 & 0.2397 & 0.0013 & 0.6635   \\
Stationary  & 0.0959 & 0.0060 & 0.7053 & 0.0041 & 0.0001 & 0.3220 & 0.0298 & 0.0052 & 0.5117 & 0.6999 & 0.0002 & 0.8201 & 0.2346 & 0.0002 & 0.6493  \\
\bottomrule
\end{tabular}}
\caption{Calculated rates by which vehicles controlled by either the \ours policy or with the baseline stationary policy are assessed to be in collision (condition C) or drive off-road (condition O), grouped by actor type and whether these outcomes were assessed to occur in the ground truth. We anchor this analysis by the provided expert demonstrations recorded in WOMD. \ours produces the expected vehicle behavior when the ground truth trajectories do not collide or drive off-road, achieving incidence rates well below one percent. While not directly rewarded in WOSAC scoring, the aversion of \ours to colliding and off-road driving is evident among circumstances where the ground truth trajectories are assessed to do so. Aside from pedestrians, all other actors controlled by \ours have incidence rates far lower than anticipated (where generated rollouts are expected to collide or drive off-road if the underlying ground truth trajectory does so). Notably, there is a proportion of trajectories that are initialized to be in collision or off-road (unsurprising given the presence of pedestrians) since the incidence rates of the stationary policy are not zero for these events in this conditional analysis.
From this, we conclude that \ours avoids an excess amount of collisions or off-road behavior as the incidence rates do not significantly increase when applying the self-play policy.}
\label{tab:wosac_collisions_and_offroad}
\end{table*}

\begin{figure*}[h]
  \centering
\includegraphics[width=\linewidth]{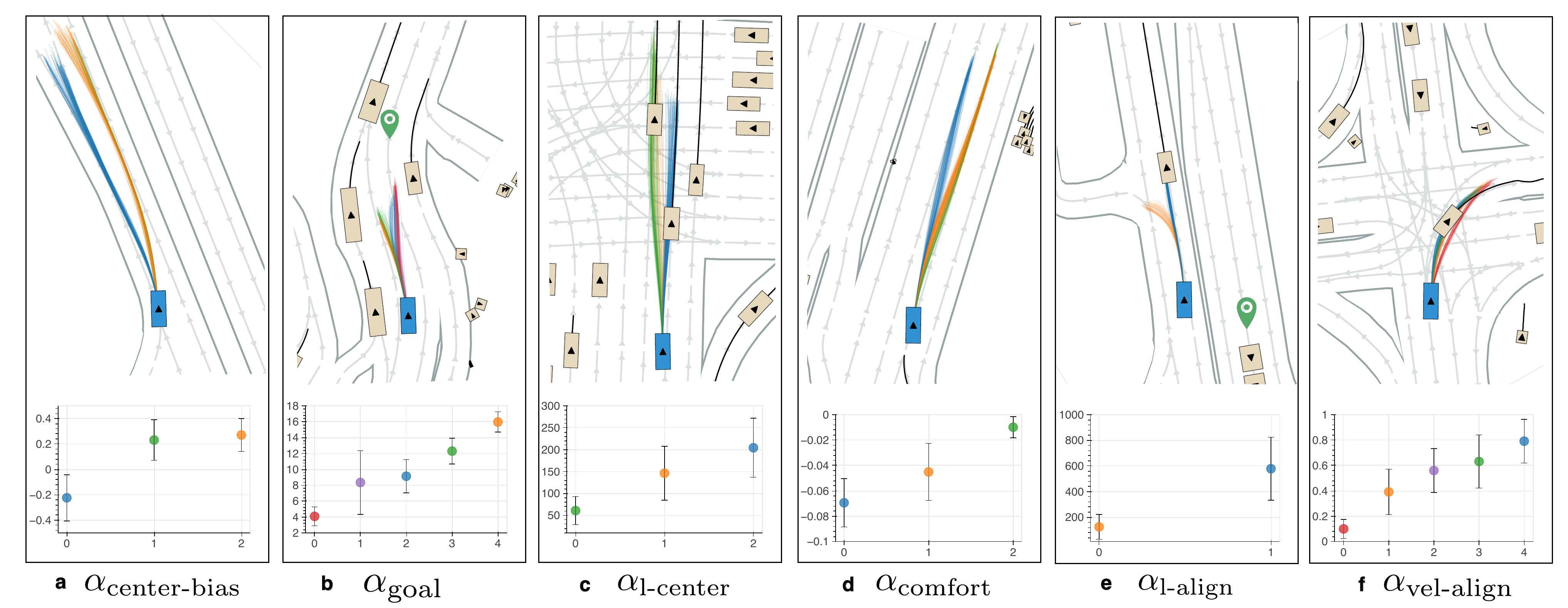}
  \caption{Conditional analysis.}
  \label{fig:conditional-analysis}
\end{figure*}

\FloatBarrier

\begin{figure*}[h]
    \centering
    \includegraphics[width=0.97\textwidth]{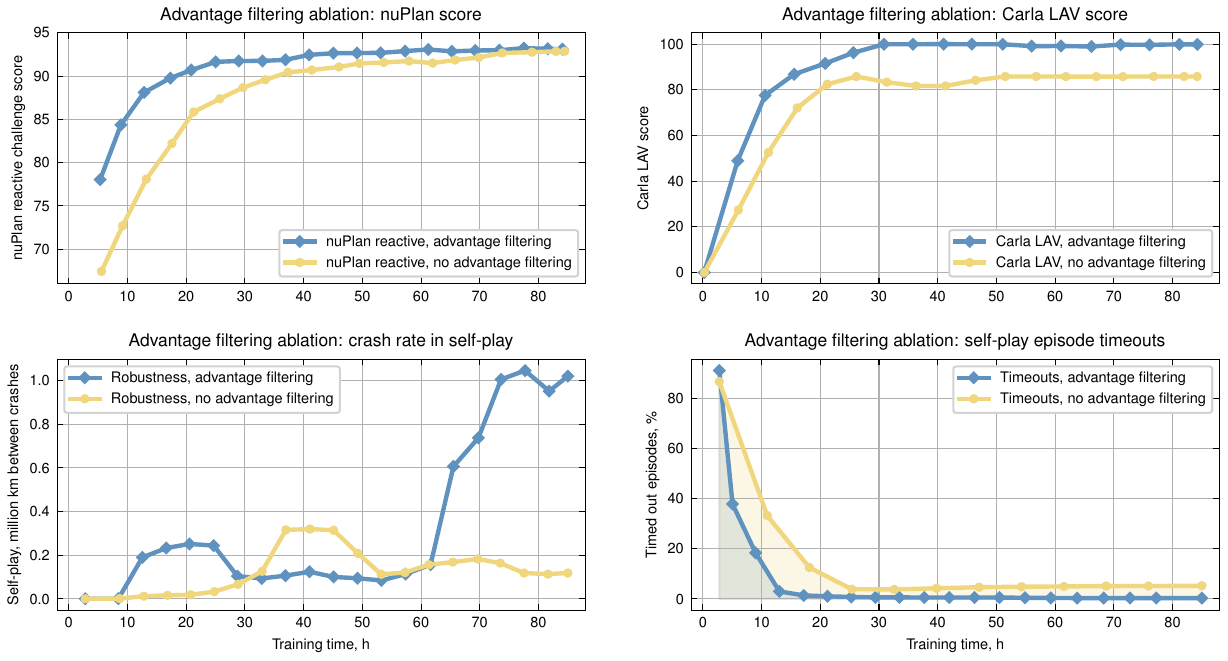}
    \caption{Training with and without \textit{advantage filtering} (see \cref{alg:advfilter}). Top row: nuPlan reactive challenge and Carla LAV Benchmark results. Bottom row: accident rate and task completion rate in self-play.}
    \label{fig:ablations_lowadv}
\end{figure*}

\begin{figure*}[h]
    \centering
    \includegraphics[width=0.75\textwidth]{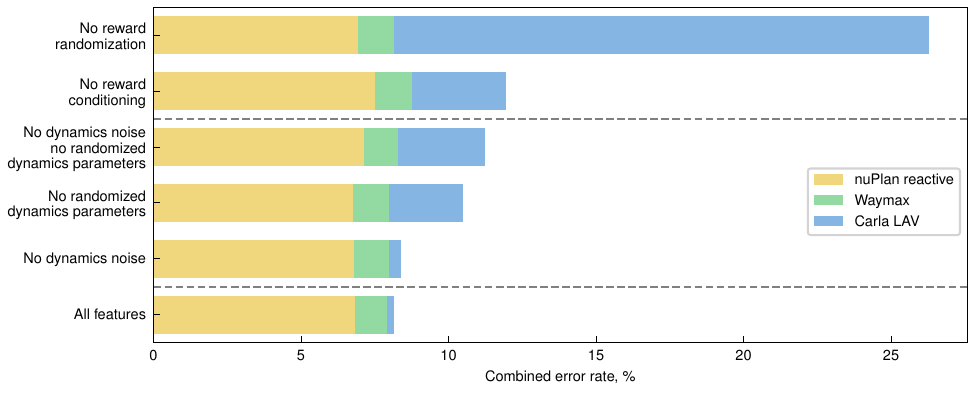}
    \caption{Aggregate impact of algorithmic features used in \ours. The length of each bar indicates the cumulative percentage points required to attain a perfect score across three benchmarks.}
    \label{fig:ablations_combined}
\end{figure*}

\begin{figure*}[h]
    \centering
    \includegraphics[width=0.65\textwidth]{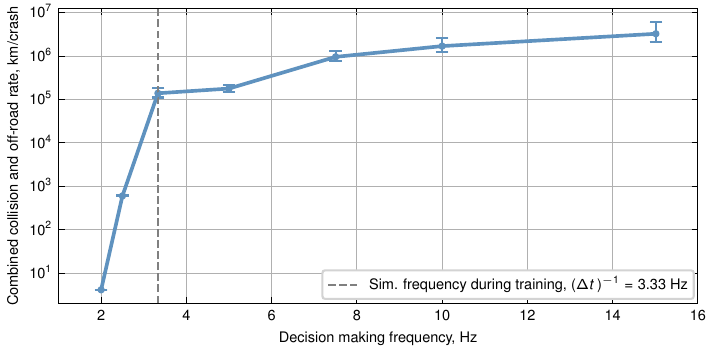}
    \caption{Accident rate in \ours self-play at different values of simulation/decision-making frequency.}
    \label{fig:self-play-robustness}
\end{figure*}

\FloatBarrier

\end{appendices}

\end{document}